\documentclass[journal]{IEEEtran}
\usepackage{cite}
\usepackage{balance}

\usepackage{algorithmic}
\usepackage{algorithm}

\usepackage[pdftex]{graphicx}
\graphicspath{{figures/}}

\usepackage{tikz}
\usepackage{tikzsymbols}
\usetikzlibrary{positioning}
\usetikzlibrary{fit}
\usetikzlibrary{decorations.pathreplacing,calligraphy}

\usepackage{url}

\usepackage{amsmath}
\usepackage{amssymb}
\usepackage{bm}
\usepackage{array}

\usepackage[caption=false,font=normalsize,labelfont=sf,textfont=sf]{subfig}
\usepackage{booktabs}

\newcommand{\aj}{\ensuremath{\mathbf{x}^a}} 
\newcommand{\ac}{\ensuremath{\mathbf{u}^a}} 
\newcommand{\ad}{\ensuremath{f^a}} 

\newcommand{\hj}{\ensuremath{\mathbf{x}^H}} 
\newcommand{\hjp}{\ensuremath{\mathbf{x'}^H}} 
\newcommand{\hjv}{\ensuremath{\mathbf{v}^H}} 
\newcommand{\hjvp}{\ensuremath{\mathbf{v'}^H}} 
\newcommand{\hc}{\ensuremath{\mathbf{u}^H}} 
\newcommand{\hd}{\ensuremath{f^H}} 
\newcommand{\hcosts}{\ensuremath{c^H}} 

\newcommand{\rj}{\ensuremath{\mathbf{x}^R}} 
\newcommand{\rc}{\ensuremath{\mathbf{u}^R}} 
\newcommand{\rd}{\ensuremath{f^R}} 
\newcommand{\rcosts}{\ensuremath{c^R}} 

\newcommand{\data}{\ensuremath{\mathcal{D}}}
\newcommand{\batch}{\ensuremath{\mathcal{B}}}

\newcommand{\hidden}{\ensuremath{\mathbf{h}}}

\newcommand{\A}{\ensuremath{\mathbf{A}}}

\usepackage{pgffor}

\begin{document}

\title{Planning Coordinated Human-Robot Motions \\
with Neural Network Full-Body Prediction Models}

\author{Philipp~Kratzer,
        Marc~Toussaint,
        and~Jim~Mainprice
        \thanks{Philipp Kratzer and Jim Mainprice are with the Machine Learning and Robotics Lab, University of Stuttgart, Germany and the Humans to Robots Motions Research Group, University of Stuttgart, Germany
        {\tt\footnotesize philipp.kratzer@ipvs.uni-stuttgart.de; jim.mainprice@ipvs.uni-stuttgart.de}}%
      \thanks{Marc Toussaint is with the Learning and Intelligent Systems lab, TU Berlin, Germany
        {\tt\footnotesize toussaint@tu-berlin.de}}
    }


\maketitle

\begin{abstract}
  Numerical optimization has become a popular approach to plan smooth motion trajectories for robots.   However, when sharing space with
  humans, balancing properly safety, comfort and efficiency still remains challenging.
  This is notably the case because humans adapt their behavior to that of the robot,
  raising the need for intricate planning and prediction. 
  In this paper, we propose a novel optimization-based motion planning algorithm,
  which generates robot motions, while simultaneously maximizing the human
  trajectory likelihood under a data-driven predictive model.  
  Considering planning and prediction together allows us to formulate
  objective and constraint functions in the joint human-robot state space.
  Key to the approach are added latent space modifiers to a differentiable human predictive model based on a
  dedicated recurrent neural network. These modifiers allow to change the human prediction within motion optimization.

   We empirically evaluate our method using the publicly available
   MoGaze \cite{kratzer2020mogaze} dataset.
   Our results indicate that
   the proposed framework outperforms current baselines for
   planning handover trajectories and avoiding collisions between a robot
   and a human.
   Our experiments demonstrate collaborative motion trajectories, where both,
   the human prediction and the robot plan,
   adapt to each other.
\end{abstract}


\IEEEpeerreviewmaketitle

\section{Introduction}

\IEEEPARstart{W}{hile} robots have been working alongside humans in factories
since the 1960's, to this day, robots are still fenced in cages and relegated to repetitive
tasks. One of the main challenges for robots to operate freely in the environment
is the difficulty to define safety and comfort objectives.
An example of a good human-robot space-sharing strategy is one that does not intervene in
the human plan while maintaining the ability for the robot to achieve its goal.
In this context the capacity to reason on some kind of predictive
human motion model is essential.

Human motion is the result of complex biomechanical processes that are challenging to model. As a consequence, state-of-the-art work on full-body motion prediction resort to data-driven models, such as recurrent neural networks (RNN)~\cite{martinez2017human, pavllo2018quaternet, wang2021pvred}. A drawback of these architectures is that they purely forecast human motion. Adapting the human prediction in order to avoid collision with the environment or plan coordinated motions, such as a handovers, is not possible.

In our prior work~\cite{kratzer2018, kratzer2020prediction},
we proposed to optimize online data-driven models by minimizing the deviation from 
the model prediction,
while accounting for environmental constraints using penalty terms.
In \cite{kratzer2020prediction} we have also shown a preliminary experiment to jointly plan a robot trajectory and predict human motion.

\begin{figure}
  \centering
  \subfloat{\includegraphics[width=0.49\columnwidth]{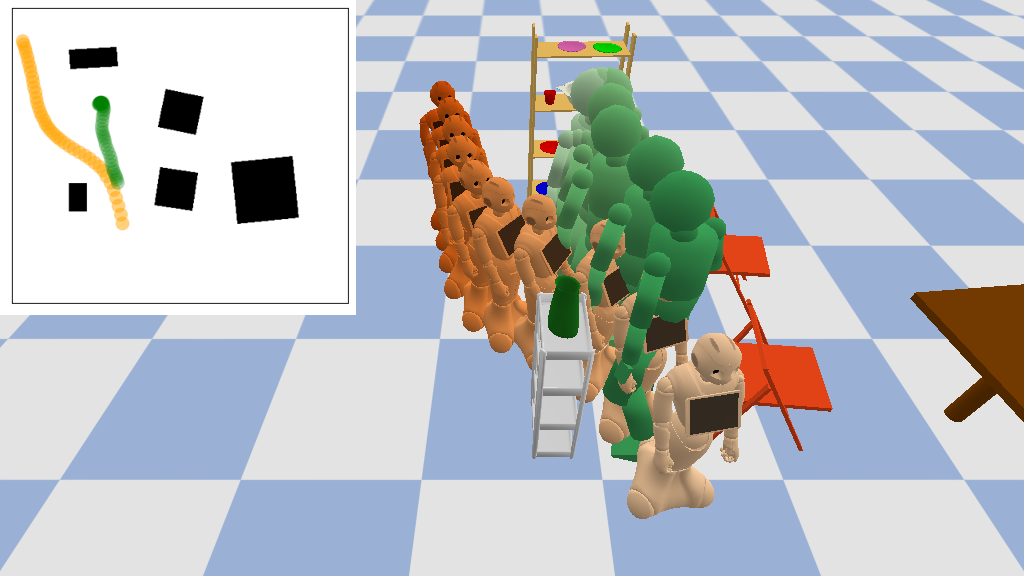}} \vspace{0.1cm}
   \subfloat{\includegraphics[width=0.49\columnwidth]{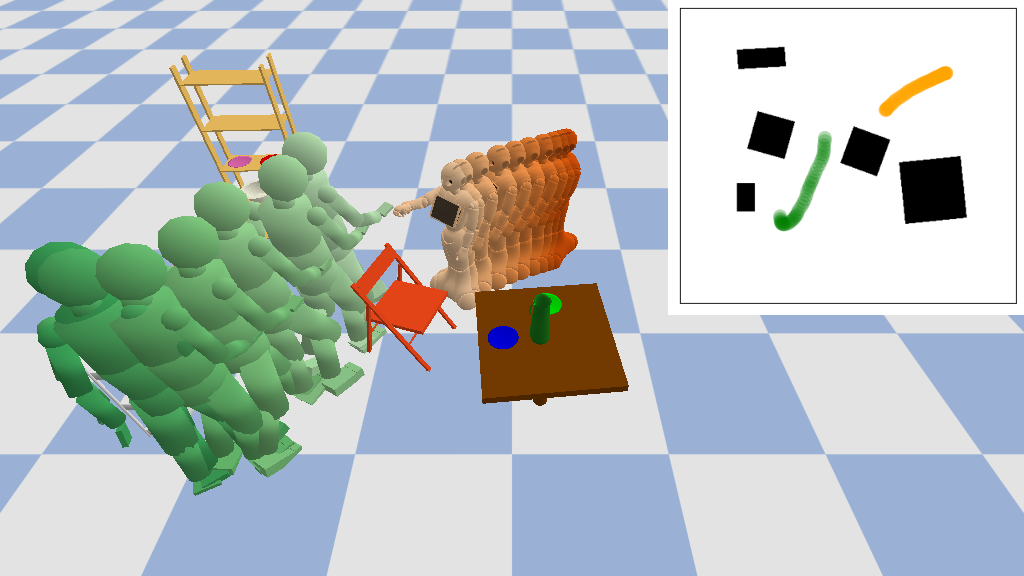}}\\
   \subfloat{\includegraphics[width=0.49\columnwidth]{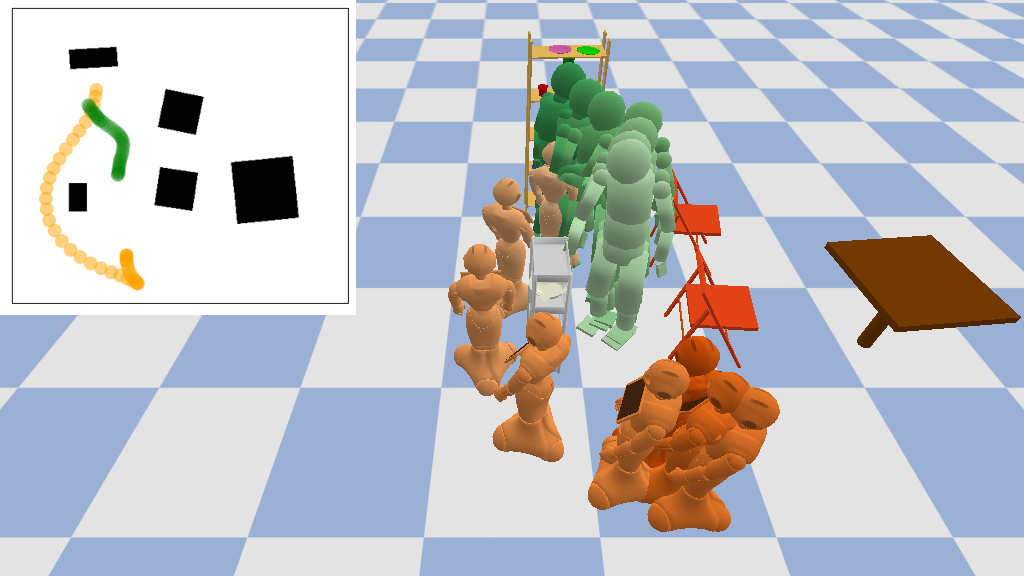}} \vspace{0.1cm}
   \subfloat{\includegraphics[width=0.49\columnwidth]{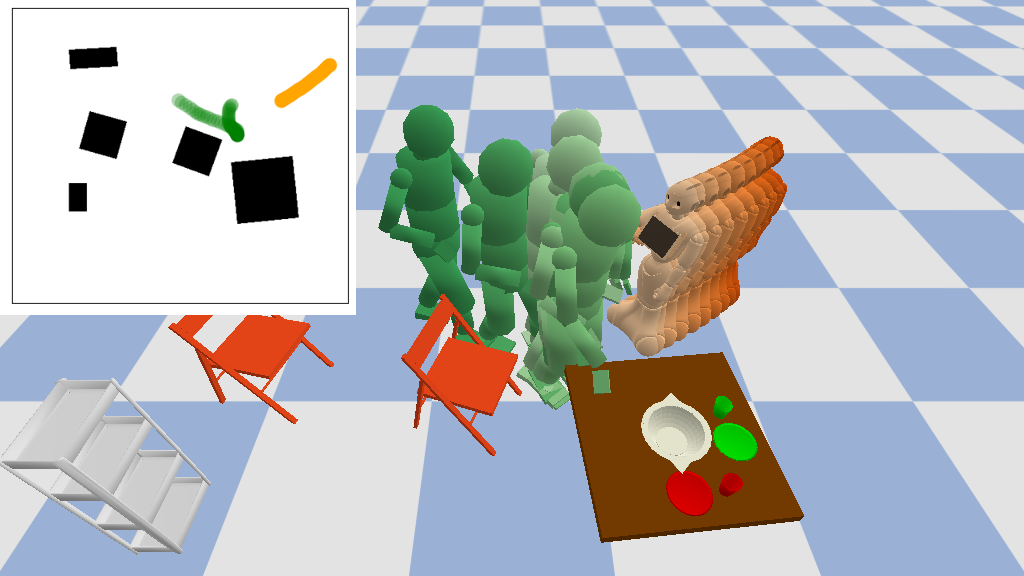}}\\
  \caption{Co-optimized Human-Robot Motion trajectories. Top Left: human and robot avoiding each other. Top Right: human and robot performing a handover. Bottom left: robot backing up to avoid human. Bottom Right: Human picking up a plate and handing it over to the robot.}
  \label{fig:intro_pic}
\end{figure}

In this work, we build on these results to propose a framework for a human-aware motion planner that uses predictive human short-term dynamics, learned by a RNN. 
In order to formulate Human-Robot Collaboration (HRC) motion planning problems, such as handovers or collision avoidance (see Figure \ref{fig:intro_pic}), as trajectory optimization, we introduce two modifications to the predictive model.

First, we adapt the architecture of the RNN by
adding latent space modifiers to the decoder inputs, leading to a
controllable dynamics function for the human. 
Second, we add differentiable Human-Robot interaction constraints 
to the output of the RNN, which take both: the robot state and the human state as input.

These modifications to the motion prediction system allow to
formulate the problem as trajectory optimization,
with decision variables for both, the robot
and the human over the entire planning horizon.
Our joint objective and constraints are entirely defined as computational graphs
including the robot model, the learned human dynamics and kinematic model,
from which we can compute gradients efficiently through automatic differentiation.

We make use of a quasi-Newton (i.e., Hessian empirical estimate) primal-dual interior-point method \cite{wachter2006implementation} to solve the corresponding nonlinear program (NLP). The result of the optimizer is a likely trajectory of human motion and a planned trajectory for the robot motion.

The main contributions of our work are:
\begin{itemize}
\item Formulation of space sharing human-robot motion planning problems formulated as 
trajectory optimization with \textit{shooting} in the joint human and robot state-space
\item Introduction of \textit{latent space modifiers} that can be used to change the human prediction.
\item Efficient gradient objective and constraints computational models
by using monolithic \textit{computational graphs}.
  
\end{itemize}

The rest of the paper is organized as follows: In Section~\ref{sec:related_work}, we discuss relevant prior work. In Section~\ref{sec:method}, we introduce our framework theoretically and explain implementation details.
In Section~\ref{sec:experiments} we evaluate our prediction framework on real motion data. We further discuss the framework in Section~\ref{sec:discussion}.
Conclusions are drawn in Section~\ref{sec:conclusions}.
\section{RELATED WORK}
\label{sec:related_work}

\subsection{Human-Aware Motion Planning}

The rapidly growing research field of HRC is focusing on robotic systems that are able to perform joint actions with humans, in order to fulfill a common task.
A main challenge in close proximity interaction
is blanching safety and comfort,
with time and energy efficient execution~\cite{lasota2017survey, ajoudani2018progress}. Pro-activity has also been investigated in many scenarios~\cite{baraglia2016initiative, schulz2018preferred}. 

In order to achieve this, the human partner needs to be taken into account
\textit{explicitly} when planning the robot's motion, leading to human-aware motion planning systems. Human-aware motion planning has been shown to improve human-robot team fluency and human worker satisfaction~\cite{lasota2015analyzing}. One way to introduce human-awareness is to incorporate a cost function to evaluate the safety of a robot path~\cite{kulic2007pre}, or predict which part of the workspace will be occupied by the human and avoid this area~\cite{Mainprice:13,lasota2014toward}. 
In order to ensure human comfort, reasoning explicitly on human’s kinematics, field of view, posture, and preferences is possible~\cite{sisbot2012human}.
For robot navigation, \textit{Proxemics}, considering
public, personal and private spaces,
are important to ensure human comfort~\cite{kruse2013human}.

In contrast to prior work in human-aware motion planning, we co-optimize robot and human motion, using a predictive human behavior model. We incorporate interaction paradigms, such as \textit{Proxemics}, as constraints in trajectory optimization.

\subsection{Human Behavior Prediction in Robotics}

In close proximity interactions with humans, 
the ability to anticipate the actions of the human partner is key.
Hence, intent prediction, which often consists of predicting a discrete action or a goal position, has been investigated in~\cite{bennewitz2005learning, elfring2014learning}. 
Object affordances can be used to improve the prediction of human intent~\cite{koppula2013, koppula2016anticipating}.
It is often required to know the full trajectory of the human. 
For example, it might be important to know which part of the workspace the human will occupy. This is often done in a second step in the aforementioned work, for example, by using social forces~\cite{elfring2014learning}.
Our method can be combined with intent prediction similarly, as we have demonstrated in our prior work~\cite{kratzer2020anticipating, le2021hierarchical}.

Many works on predictive behavior models focus on directly forecasting human motion. While 2d human motion prediction is especially important for robot navigation~\cite{rudenko2020human, ziebart2009planning, gupta2018social}, we are not only interested in modeling navigation, but also in pick and place tasks or handovers and thus require a full-body motion prediction model.

\subsection{Human Full-Body Prediction Models}
\label{ssec:fbpred}
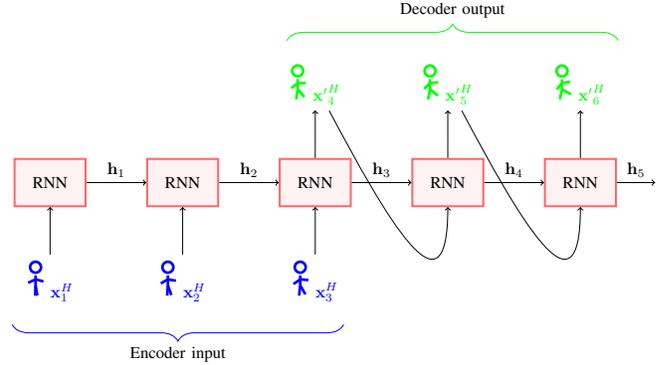
\begin{figure}
  \centering
  \resizebox{\columnwidth}{!}{%
      \begin{tikzpicture}[
roundnode/.style={circle, draw=black, fill=black!5, very thick, minimum size=7mm},
squarednode/.style={rectangle, draw=red!60, fill=red!5, very thick, minimum size=10mm, minimum width=15mm},
]
\foreach \x in {1,2,3} {
\node[squarednode]        (node\x) at (2.8*\x, 0)    {RNN};
\node[text=blue] (nodestrich\x) [below=of node\x] {\Strichmaxerl[3][20+5*\x][-20+5*\x][-2*\x][10*\x]$\hj_\x$};
\draw[->] (nodestrich\x) -- (node\x);
}
\foreach \x in {4,5} {
  \node[squarednode] (node\x) at (2.8*\x, 0)    {RNN};
}
\foreach \x in {4,5, 6} {
  \pgfmathtruncatemacro{\prev}{subtract(\x,1)}
  \node[text=green] (nodestrich\x) [above=of node\prev] {\Strichmaxerl[3][20+5*\x][-20+5*\x][-2*\x][10*\x]$\hjp_\x$};
  \draw[->] (node\prev) -- (nodestrich\x);
}

\node[]        (node6) at (15.7, 0)    {};

\foreach \step in {1,...,5} {
  \pgfmathtruncatemacro{\next}{add(\step,1)}
  \draw[->] (node\step) -> (node\next) node[midway,above] {$\hidden_\step$};
}

\foreach \step in {4,5} {
  \draw[->] (nodestrich\step) to [out=298, in=270, looseness=2.] (node\step);
}
\draw [decorate, decoration = {brace, amplitude=10pt}, blue] (9,-3) --  (2,-3) node [black,midway, below,yshift=-10pt] {Encoder input};
\draw [decorate, decoration = {brace, amplitude=10pt}, green] (7.8,3) --  (14.8,3) node [black,midway, above,yshift=10pt] {Decoder output};
\end{tikzpicture}
}
\caption{Human Motion Prediction with a RNN. The observed trajectory (blue) is fed into RNN cells (red) and future states can  be predicted (green).}
\label{fig:rnn_pred}
\vspace{-.5cm}
\end{figure}

Early methods for full-body or arm prediction, for example, use inverse optimal control~\cite{berret2011evidence, mainprice2016goal}. However, the availability of larger human motion data-sets and recent advances in neural networks make deep learning techniques state-of-the-art. Due to the sequential structure of motion data, RNNs are suitable for full-body motion prediction~\cite{fragkiadaki2015recurrent}. A typical encoder-decoder structure can be seen in Figure~\ref{fig:rnn_pred}. The architecture can be further improved by adding residual connection in the loop function~\cite{martinez2017human}. It has also been shown that the rotation representation and loss is important, for instance, using a quaternion representation improved over prior work~\cite{pavllo2018quaternet, pavllo2019modeling}. Including a velocity connection can make predictions more stable for longer time horizons~\cite{wang2021pvred}.
Recently, motion prediction using graph neural networks~\cite{li2020dynamic} or transformers~\cite{aksan2021spatio} has been shown to slightly improve the prediction performances.

Motion prediction based on neural networks promises good results for predicting short-term motion. However, the models have the issues that 1) they purely forecast human motion and do not incorporate workspace geometry 2) they are not controllable and thus can not be changed during motion planning. In our work we tackle these issues by adding modifiers to the network architecture, which allows for optimization-based motion prediction.

\begin{figure*}
\centering
\includegraphics[width=\linewidth]{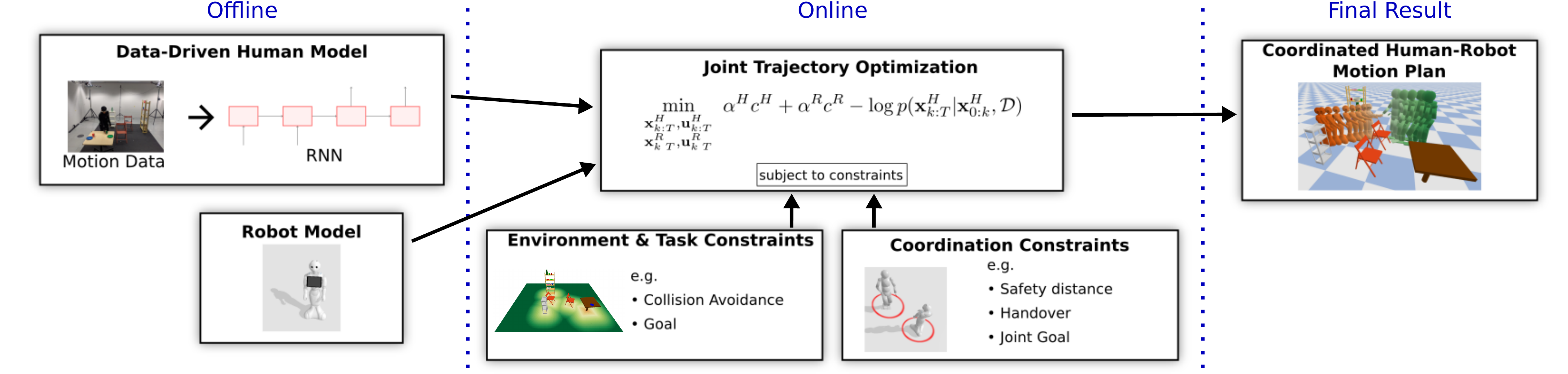}
\caption{Overview of the framework.  Offline a human model is trained on motion data and a robot model is constructed. Online the states and controls of human and robot are jointly optimized, accounting for task, environment and, coordination constraints. The final result is a coordinated motion plan.}
\label{fig:overview}
\vspace{-.3cm}
\end{figure*}

\subsection{Motion Optimization}
\label{ssec:motion_optim}

Gradient-based optimization algorithms are widely used in the field of robotics and optimal control  for optimizing trajectories~\cite{todorov2005generalized, ratliff2009chomp, schulman2013finding, toussaint2014newton, marinho2016functional, Toussaint:17, mainprice2020interior}.
These techniques have been shown to successfully generate motions with
a variety of kinematic and dynamic objectives and constraints, such as obstacle constraints~\cite{ratliff2009chomp}.

Motion optimization has also been used to synthesize human behavior for animating characters and is able to generate realistic motions~\cite{mordatch2012discovery, mordatch2013animating}. In contrast, our work focuses on forecasting human motion using a data-driven model and uses motion optimization for planning robot motion while adapting the human prediction to the robot plan.

Since we first proposed this idea in~\cite{kratzer2020prediction}, similar ideas have been proposed in the meantime. 

For example, Fishman et al. propose a method to simultaneously plan for the robot arm movements while predicting human actions a floating sphere model~\cite{fishman2020collaborative}.
Schaefer et al. use gradient information of a neural network for planning 2D trajectories within a crowd of pedestrians~\cite{schaefer2021leveraging}. 

In contrast to those works, our approach makes use of a full-body motion prediction model for the human, which makes it possible to plan motions using a single framework for tasks such as handovers, or tasks where navigation and pick and place actions are combined.

\section{Joint Human-Robot Trajectory Optimization and 
Motion Prediction}
\label{sec:method}

\subsection{Problem Statement}
\subsubsection{Agents}
We consider scenarios where multiple agents $a$ are involved, who have to solve specific tasks in the environment. To simplify, we consider and evaluate our framework with one human agent~$H$ and one robot agent~$R$.

We discretize time with a small constant time change of $\Delta t$ between consecutive timesteps. At a timestep $t$, each agent $a \in \{H,R\}$ has a state $\aj_{t}\in\mathbb{R}^{d_a}$ with dimensionality $d_a$. When applying controls $\ac_{t}\in\mathbb{R}^{n_a}$ a discrete dynamics function $\ad$ can be used to compute the state at the next timestep:
\begin{align}
  \aj_{t+1} = \ad ( \aj_{t}, \ac_{t})
\end{align}

\subsubsection{Formulation as NLP}

We formulate our HRC problem, which aims to plan a robot trajectory $\rj_{k:T}$ while predicting a likely human trajectory $\hj_{k:T}$, as the following NLP:
\begin{align}
  \min_{\substack{\hj_{k:T},\hc_{k:T}\\\rj_{k:T},\rc_{k:T}}}~ &\alpha^H \hcosts + \alpha^R \rcosts - \log p(\hj_{k:T}|\hj_{0:k}, \data)\label{equ:hrnlp}\\
  \text{subject to:}& \notag\\
  &\aj_{t+1} = \ad(\aj_t, \ac_t)~\forall t \in (k,T) \label{equ:hdc}\\
  &h^a(\aj) = 0,~ g^a(\aj) \leq 0 \label{equ:heic}\\
  &h(\hj, \rj) = 0,~  g(\hj, \rj) \leq 0 \label{equ:hreic}
\end{align}
with $\hcosts(\hj_{k:T}, \hc_{k:T})$ and $\rcosts_t(\rj_{k:T}, \rc_{k:T})$ being cost functions associated with the trajectories of human and robot, $\alpha^H$ and $\alpha^R$ are hyperparameters used to weight the influence of the respective agents,
and the likelihood of the human motion given
dataset $\data$ and observed motion $\hj_{0:k}$ is given by $p(\hj_{k:T}|\hj_{0:k}, \data)$.

Equation~(\ref{equ:hdc}) shows constraints ensuring the dynamics functions for human and robot. Additional equality and inequality constraints for human or robot (Equation~\ref{equ:heic}) can be used to ensure environment-dependent and task constraints. Joint constraints between human and robot (Equation~\ref{equ:hreic}) are useful to ensure collaborative interaction paradigms.

\subsubsection{Trajectory optimization with shooting}

One possibility to solve the trajectory optimization problem is a \textit{collocation} approach,
which means having both, the controls and the states as part of the decision variables and specifying the dynamics as explicit constraints. However, since the recurrent neural network introduces additional hidden states $\hidden_{0:T}$ per timestep, those would also need to be specified as equality constraints, leading to an enormous number of constraints and high memory requirements.

Nevertheless, when future controls $\ac_{t:T}$ are available, the future states $\aj_{t+1:T}$ can be computed by forward simulation using the unrolled dynamics $\tilde\ad$, which applies $\ad$ recursively:
\begin{align}
\aj_{t+1:T} &= \tilde \ad(\aj_{t}, \ac_{t:T}) \notag\\
&=\left(\ad(\aj_t, \ac_t),~ \ad(\ad(\aj_t, \ac_t), \ac_{t+1}),~ \ldots\right)
\end{align}

This is the concept underlying the \textit{shooting} approach to trajectory optimization.
As a consequence, dynamic constraints are directly fulfilled and only controls,
$\ac_{k:T}$ are part of the decision variables. 
A disadvantage of this approach is that the controls at a specific timestep $\ac_t$ have an influence on all states at later timesteps $\aj_{t+1:T}$. As a consequence, it is required to propagate the gradient through the unrolled dynamics for optimization.

\subsection{Solving the NLP}
Using the shooting method and the assumption that the human likelihood is integrated into the human dynamics function (see Section~\ref{sssec:uhd}), the  optimization problem simplifies to:
\begin{align}
  \min_{\substack{\hc_{k:T},\rc_{k:T}}}~ &\alpha^H \hcosts + \alpha^R \rcosts\\
  \text{subject to:}& \notag\\
  &h^a(\aj) = 0,~ g^a(\aj) \leq 0 \\
  &h(\hj, \rj) = 0,~  g(\hj, \rj) \leq 0
\end{align}
\noindent
where $c$ are the costs and $h$ and $g$ are user defined equality and inequality constraints explained in Section~\ref{ssec:constraints}. Note that the constraints can act on any state $\hj_t$ or $\rj_t$ at any prediction timestep $t$. Thus, the unrolled dynamics~$\tilde \hd$ of the neural network and its derivatives are required.

\subsubsection{Primal-Dual Interior Point Method for Non-linear programming}

Gradient-based methods can be used to optimize the problem with, for example, specifying constraints using barrier functions or by using Lagrange multipliers. We use an interior-point gradient-based solver (ipopt~\cite{wachter2006implementation}) with linear solver MA57~\cite{duff2004ma57} for optimizing the NLP. 

Interior point methods use barrier functions to convert the constrained optimization problem into a sequence of unconstrained barrier problems:
\begin{align*}
  B(x,\mu) = c(x) - \mu  \sum_{i=1}^n  \ln(g_i(x))
\end{align*}
where $c$ is the objective function, $g_i$ are the inequality constraints and $\mu$ being the barrier parameter.
Ipopt computes solutions to the barrier problems, and then decreases the barrier parameters $\mu$ to drive them towards 0. As we do not provide the analytic Hessian, ipopt is approximating the Hessian using a BFGS update.

We use the tensorflow library~\cite{abadi2016tensorflow} to formulate the constraints end-to-end, directly in the network architecture (see Figure~\ref{fig:rnn_coll_constr}), which allows us to use the automatic differentiation functionality to obtain the gradients.
Algorithm~\ref{alg:joint_trajopt} shows a high-level overview over the optimization procedure.

\begin{algorithm}[t]
\caption{Human-Robot Joint Trajectory Optimization}
\begin{algorithmic}
\STATE
\STATE \textbf{Input:} Observed trajectory $\hj_{0:k}$, Optimization parameters $\alpha^R, \alpha^H$
\STATE \textbf{Init:} $\hc_{k:T}\gets\mathbf{0},~\rc_{k:T}\gets\mathbf{0}$
\FOR{$n~<$ maxiter}
\STATE $\hj_{k:T} \gets \tilde \hd(\hj_{k}, \hc_{k:T})$
\STATE $\rj_{k:T} \gets \tilde \rd(\rj_{k}, \rc_{k:T})$
\STATE Compute objectives $\hcosts,~\rcosts$ and constraints $h_i,~g_i$
\STATE Compute gradients $\frac{\partial \hcosts}{\partial \hc}$, $\frac{\partial \rcosts}{\partial \rc}, \frac{\partial h_i}{\partial \hc}$, $\frac{\partial h_i}{\partial \rc},\frac{\partial g_i}{\partial \hc}$, $\frac{\partial g_i}{\partial \rc}$ (tensorflow)
\STATE Compute logbarrier parameters (ipopt)
\STATE Approximate Hessian with BFGS (ipopt)
\STATE Compute search direction and step size (ipopt)
\STATE Update controls $\hc_{k:T},~\rc_{k:T}$
\ENDFOR
\RETURN $\hj_{k:T},~\hc_{k:T},~\rj_{k:T},~\rc_{k:T}$
\end{algorithmic}
\label{alg:joint_trajopt}
\end{algorithm}

\subsection{Motion Objectives and Constraints}
\label{ssec:constraints}

In this section we show examples of motion constraints, which we found to work well in our experiments.

Constraints $h$ are equality constraints and $g$ inequality constraints. The differentiable kinematics function $\phi^a$ is used to calculate positions or orientations attached to a specific link of the agent.

\subsubsection{Objective}

We penalize the magnitude of the control terms
$\hcosts=||\hc||^2_\A$ and $\rcosts=||\rc||^2_\A$, where $\A$ is a norm
which approximates $\hcosts \approx \sum_t^T ||\dot{\mathbf{u}}_t^a||^2$ by finite difference.

Since the neural network is trained to find a likely prediction of human motion (See section \ref{sssec:uhd}) and thus approximates $p(\hj_{k:T}|\hj_{0:k}, \data)$, minimizing $\hcosts=||\hc||^2_\A$ pulls the neural network cell inputs closer to the inputs of the initial prediction. 
Hence we assume that minimizing $\hcosts$ leads to a similar effect as maximizing the likelihood of future states. Note that setting $\hc=\mathbf{0}$ leads to the initial prediction of the neural network, which we use to warm start the optimizer.

\subsubsection{Single agent constraints}
\label{ssec:saconstr}

Often we want to plan trajectories where an agent needs to move to a specific position or needs to pick up a specific object. Thus, we formulate a \textit{goal constraint} as equality constraint with a specific link, such as hand or base, ending up at a specific goal position $p^*$ at timestep $t$:
\begin{align}
h_t^\text{Goal} = ||
(\phi_\text{linkpos}^a \circ \tilde f_{t}^a) (\aj_{k}, \ac_{k:t}) - p^* ||^2
\end{align}
where $\phi_\text{linkpos}^a(\aj)$ is the forward kinematics (FK) of a given link for agent $a$.

In order to avoid collision with obstacles in the environment at timestep $t$, we use a Signed Distance Field (SDF) and formulate the \textit{collision constraint} as follows:

\begin{align}
g_t^\text{Collision} = 
(\text{SDF} \circ \phi_\text{basepos}^a \circ f_{t}^a)(\aj_{k}, \ac_{k:t})
\end{align}
where $\phi_\text{basepos}^a$ is the base position of $a$.

For avoiding collision at all timesteps, one constraint for every timestep can be defined. However, to save computation time, we use a single constraint using a max function as follows:
\begin{align}
  g^\text{Collision} = \text{max}(g_k^\text{Collision}, g_{k+1}^\text{Collision}, \ldots, g_T^\text{Collision})
\end{align}
We found that both approaches work reasonably well. In our implementation we specify the
collision constraint as a single constraint because it allows us to compute all needed $g_t$ while unrolling $\tilde f$ and using a single automatic differentiation call to compute the gradient.

Figure~\ref{fig:rnn_coll_constr} shows the integration of the constraint into the RNN architecture depicted in gray. 
Note that to improve the optimization behavior, it can make sense to use a smooth approximation of the maximum function, such as a RealSoftMax (a.k.a. LogSumExp).

\subsubsection{Joint constraints}
\label{ssec:jconstr}

To avoid collision between the agents, we want to maintain a minimal clearance of $d$ between the base positions of the agents at all time. We formulate this as an inequality constraint with the following~\textit{joint collision constraint} at every timestep $t$:
{\small
\begin{align}
  &g^\text{Jointcoll}_t = \notag\\
  &~~~||
  (\phi_\text{basepos}^H \circ f_{T}^H) (\hj_{t}, \hc_{k:t}) - 
  (\phi_\text{basepos}^R \circ f_{t}^R) (\rj_{t}, \rc_{k:T})-d||^2 \label{equ:jointcoll}
\end{align}}

Similar to $g^\text{Collision}$ it is possible to formulate it as a single constraint using the maximum over $g^\text{Jointcoll}_{k:T}$.

In case we want to pickup an object, but leave to the optimizer,
which of the agents should pick up the object and when the pickup should happen,
the following \textit{joint goal constraint} per timestep $t$ can be used:
\begin{align}
  h^\text{Jointgoal}_t = \text{min}(&||
  (\phi_\text{hand}^H \circ f_{T}^H) (\hj_{t}, \hc_{k:t}) - p^*||^2, \label{equ:jointgoal}\\
&||( \phi_\text{hand}^R  \circ f_{t}^R)(\rj_{t}, \rc_{k:T}) - p^*||^2)\label{equ:jointgoal2}
\end{align}
with $p^*$ being the grasp point for the object. To compute it over all the prediction timestep we use the minimum over $h^\text{Jointgoal}_{k:T}$ letting the optimizer decide, which timestep is suitable for the grasp.

In order to plan for handover motion we consider the following \textit{Handover constraint}:
\begin{align}
&h^\text{Handover} =\notag\\
&~~~||
(\phi_\text{handpos}^H \circ f_{T}^H) (\hj_{t}, \hc_{t:T}) - 
(\phi_\text{handpos}^R \circ f_{T}^R) (\rj_{t}, \rc_{t:T})||^2 + \label{equ:handover_handpos}\\
&~~~||
(\phi_\text{baserot}^H \circ f_{T}^H) (\hj_{t}, \hc_{t:T}) - 
(\phi_\text{baserot}^R \circ f_{T}^R) (\rj_{t}, \rc_{t:T})||^2\label{equ:handover_baserot}
\end{align}
where the first part (Equation~\ref{equ:handover_handpos}) of the handover loss is the distance between the hand positions of both agents. We define $\phi_\text{handpos}$ to be an offset to the hand palms so that the hands end up in a distance suitable for handing over objects. The second part (Equation~\ref{equ:handover_baserot}) is the distance of the base rotations so that it equals to 0 when the agents face each other.

\subsection{Agent Dynamics Functions}
\label{ssec:adfun}
\subsubsection{Robot Dynamics}

A typical dynamics function for the robot agent $\rd$
uses torques as controls $\rc$ and position and velocities as states $\rj$.
Hence the dynamics function can be derived from the connectivity of the kinematic chains
and the mass parameters of the robot.

In our experiments 
we use a Pepper\footnote{\url{https://www.softbankrobotics.com/emea/en/pepper}}
robot with active joints for its right arm.
We model the dynamics as simple velocity control,
where the states is composed of the robot 2D base position $\rj_{t, 0:2}$ and orientation $\rj_{t,2}$. 
The controls are velocity $\rc_{t,0}$ and angular velocity $\rc_{t,1}$. This leads to the following dynamics function:
\begin{align*}
\rj_{t+1} = \rd(\rj_t, \rc_t) =   \left(
\begin{array}{c}
\rj_{t, 0} + \cos(\rj_{t, 2}) \cdot \rc_{t,0}\\
\rj_{t, 1} + \sin(\rj_{t, 2}) \cdot \rc_{t,0}\\
  \rj_{t, 2} + \rc_{t,1}\\
  \ldots \\
  \rj_{t, d_R} + \rc_{t,d_R-1}\\
\end{array}
\right) 
\end{align*}

\noindent
where $\rj_{t, 3:d_R}$ are other velocity controlled state dimensions that arise from the robot kinematics.

\subsubsection{Uncontrolled Human Dynamics using a RNN}
\label{sssec:uhd}
We model the kinematic state of a human as a vector consisting of base position, base rotation and joint angles: $\hj = (p_{\text{base}}, r_{\text{base}}, r_{\text{joints}})$.
We then approximate the likelihood of the human motion $p(\hj_{k:T}|\hj_{0:k}, \data)$
using a RNN similar to the one presented in~\cite{wang2021pvred} by training it on the demonstration data $\data$. 
This leads to an uncontrolled predictive model of human movement.

\begin{figure}
  \centering
  \resizebox{\columnwidth}{!}{%
      \begin{tikzpicture}[
  roundnode/.style={circle, draw=black, very thick, minimum size=2mm, inner sep=0pt},
  smalldot/.style={circle, draw=black, fill=black, minimum size=1mm, inner sep=0pt},
  deltas/.style={circle, draw=orange, fill=orange!5, minimum size=5mm, inner sep=0pt, minimum width=5mm, thick},
  squarednode/.style={rectangle, draw=red!60, fill=red!5, thick, minimum size=5mm, minimum width=15mm},
  greennode/.style={rectangle, draw=green!60, fill=green!5, thick, minimum size=5mm, minimum width=15mm},
    constrainnode/.style={rectangle, draw=gray!60, fill=gray!5, thick, minimum size=5mm, minimum width=15mm},
  minimumnode/.style={rectangle, draw=gray!60, fill=gray!5, thick, minimum size=5mm, minimum width=70mm},
  bluenode/.style={rectangle, draw=blue!60, fill=blue!5, thick, minimum size=5mm, minimum width=15mm},
]
\foreach \x in {1,2,3} {
  \node[squarednode]        (nodel3\x) at (2.8*\x, 0)    {GRU};
  \node[squarednode]        (nodel2\x) at (2.8*\x, .7)    {GRU};
  \node[squarednode]        (nodel1\x) at (2.8*\x, 1.4)    {Linear};
  \node[bluenode] (nodeRotrep\x) at (2.8*\x, -1.8) {Rot. Rep.};

  \node[fit=(nodel1\x)(nodel2\x)(nodel3\x), draw,rounded corners=.1cm] (node\x){};
\node[text=blue] (nodestrich\x) at (2.8*\x, -3.2) {\Strichmaxerl[3][20+5*\x][-20+5*\x][-2*\x][10*\x]$\hj_\x$};
\node[roundnode] (concatnode\x) at (2.8*\x, -.9) {};
\node[] (velocitynode\x) at (2.8*\x-0.8, -.9) {$\hjv_\x$};

\draw[->] (nodestrich\x) -- (nodeRotrep\x);
\draw[->] (nodeRotrep\x) -- (concatnode\x);
\draw[->] (concatnode\x) -- (node\x);
\draw[->] (velocitynode\x) -- (concatnode\x);
}
\foreach \x in {4,5} {
  \node[squarednode]        (nodel3\x) at (2.8*\x, 0)    {GRU};
  \node[squarednode]        (nodel2\x) at (2.8*\x, .7)    {GRU};
  \node[squarednode]        (nodel1\x) at (2.8*\x, 1.4)    {Linear};
  \node[fit=(nodel1\x)(nodel2\x)(nodel3\x), draw,rounded corners=.1cm] (node\x){};
  \node[smalldot] (nodedot\x) at (2.8*\x+1.4-2.8, 2.4) {};
  \node[roundnode] (concatnode\x) at (2.8*\x, -.9) {};
  \node[smalldot] (velnodedot\x) at (2.8*\x-2.8, 2.) {};
    \node[smalldot, color=orange, minimum size=2mm] (deltaveldot\x) at (2.8*\x-.5, -1.3) {};
  \node[smalldot, color=orange, minimum size=2mm] (deltaposdot\x) at (2.8*\x-.5, -.9) {};
  \node[deltas] (nodedelta\x) at (2.8*\x-.5, -2) {\hc};
  \draw[color=orange, line width=0.5mm] (nodedelta\x) -> (deltaposdot\x);
}

\node[minimumnode] (nodeMinimum) at (2.8*4, 7.8) {Minimum};
\node[] (aboveminimumnode) at (2.8*4, 8.5) {};
\draw[->] (nodeMinimum) -- (aboveminimumnode);

\foreach \x in {4,5, 6} {
  \pgfmathtruncatemacro{\prev}{subtract(\x,1)}
      \node[roundnode] (nodeplus\x) at (2.8*\prev, 2.4) {+};
      \node[text=green] (nodestrich\x) at (2.8*\prev, 4.6) {\Strichmaxerl[3][20+5*\x][-20+5*\x][-2*\x][10*\x]$\hjp_\x$};
      \node[greennode] (nodeRotrep\x) at (2.8*\prev, 3.4) {Rot. Rep.};

      \node[constrainnode] (nodeFK\x) at (2.8*\prev, 5.8) {FK};
      \node[constrainnode] (nodeSDF\x) at (2.8*\prev, 6.8) {SDF};
      \draw[->] (nodeplus\x) -- (nodeRotrep\x);
      \draw[->] (nodeRotrep\x) -- (nodestrich\x);
      \draw[->] (nodestrich\x) -- (nodeFK\x);
      \draw[->] (nodeFK\x) -- (nodeSDF\x);
      \draw[->] (nodeSDF\x) -- (nodeMinimum);
      }

\foreach \x in {4,5} {
  \pgfmathtruncatemacro{\prev}{subtract(\x,1)}
  \draw[->] (node\prev) -- (nodeplus\x);
}
  \draw[->] (node5) -- (nodeplus6) node[near end, right] {\footnotesize $\hjvp_6$};

\node[]        (node6) at (15.7, .7)    {};

\foreach \step in {1,2, 5} {
  \pgfmathtruncatemacro{\next}{add(\step,1)}
  \draw[->] (node\step) -> (node\next) node[midway,above] {$\hidden_\step$};
}
\foreach \step in {3,4} {
  \pgfmathtruncatemacro{\next}{add(\step,1)}
  \draw[->] (node\step) -> (node\next) node[near end,above] {$\hidden_\step$};
}
  \node[smalldot] (nodedot3) at (2.8*3, -1.2) {};
\draw[->] (nodedot3) -| ++(-1.1,0) |- (nodeplus4);
\foreach \step in {4, 5} {
  \pgfmathtruncatemacro{\next}{add(\step,1)}
  \draw[->] (nodedot\step) -> (nodeplus\next);
}
\foreach \step in {4,5} {
  \node[smalldot] (loopbackdot\step) at (2.8*\step -2.8, 2.7) {};
  \draw[->] (loopbackdot\step) -| ++(1.4,0) |- ++(0.,-4.) |- ++(1.4,0.) -| (concatnode\step.south);
  \draw[->] (velnodedot\step) -| ++(1.1,0) node[near start, above] {\footnotesize $\hjvp_\step$} |- ++(0.,-2.9) |- ++(1.4,0.) |- (concatnode\step.west);
  \draw[->] (concatnode\step) -- (node\step);
}
\end{tikzpicture}

}
\caption{Network Architecture. Joint states (blue) serve as input to the RNN cells (red), which output the predictions (green). Controls (orange) can be used to change the encoder inputs.  For example, a collision constraint can use  FK and SDF layers (gray).}
\label{fig:rnn_coll_constr}
\end{figure}
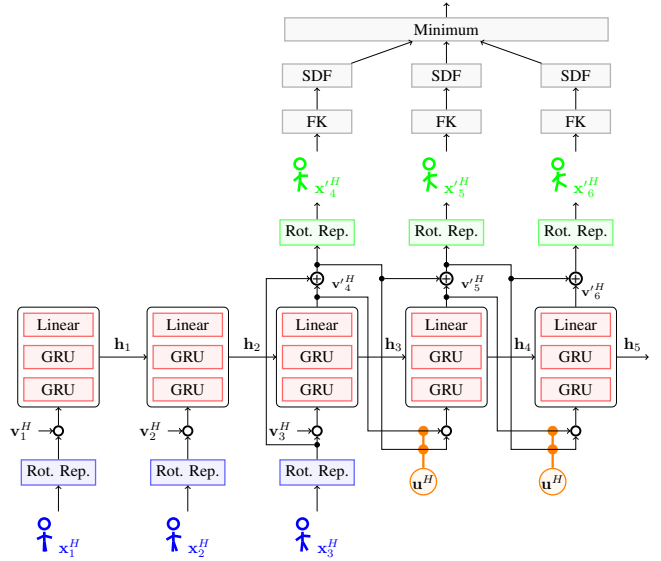

An example RNN architecture with only 6 timestep can be seen in Figure~\ref{fig:rnn_coll_constr}. The ground truth joint states $\hj_{0:3}$ serve as input to the neural network (depicted in blue). In contrast to most prior work in motion prediction we convert the rotation representation to a continuous 6D representation~\cite{zhou2019continuity} instead of using an axis angle representation or quaternions. The state is augmented with velocities $\hjv$, which we compute using finite differences $\hjv_t=\hj_{t+1}-\hj_t$. 

The concatenated state serves as input to a stack of $l$ gated recurrent units (GRU) layers and one linear layer (depicted in red). We do not feed the position of the base $p_\text{base}$ into the recurrent unit. This avoids conditioning the model on world positions and leads to better generalization. The stacked layers output the predicted velocities $\hjvp$, which is achieved by adding the previous state to the velocities by using a residual connection. 

In the decoder part, the outputs of the previous cell are used as input to the following cell, which allows to predict for multiple timesteps. Converting back the rotation representations results in the final predicted states $\hjp_{4:6}$ (depicted in green).

The function $\hd$ obtained from the RNN is the uncontrolled dynamics function:
\begin{align}
  (\hj_{t+1},\hjv_{t+1}, \hidden_{t+1})  = \text{RNNCell}_{\mathbf{\theta}} (\hj_{t},~\hjv_{t},~\hidden_t) \label{equ:ucdyn}
\end{align}
with the state being augmented with velocities $\hjv$ and the hidden state $\hidden$ of the RNN and with $\mathbf{\theta}$ being the network parameters.

\subsubsection{Controlled Human Dynamics}
With Equation~(\ref{equ:ucdyn}) we are able to produce a likely prediction for future human states. However, this prediction is uncontrolled and we can not change it when we plan ahead.

In order to change the prediction, we add latent space modifiers $\hc$ to the inputs of the decoder RNN cells (see Figure~\ref{fig:rnn_coll_constr}, depicted orange). We modify both, the input state $\hj_t$ by adding the modifiers $\hc_t$ and the input velocities $\hjv_t$ by adding finite differences of the modifiers $\hc_{t+1}-\hc_t$ at each prediction timestep. This changes the uncontrolled dynamics (Equation~(\ref{equ:ucdyn})) to the controlled version:
\begin{align}
  \hj_{t+1} &= \hd(\hj_t, \hc_t) \notag\\
  &= \text{RNNCell}_{\bm{\theta}}(\hj_t + \hc_t,~\hjv_t + \hc_{t+1} - \hc_t,~\hidden_t)
\end{align}
We can now change the state and velocity input at every predicted timestep by using the latent space modifiers~$\hc$. While the latent space modifiers $\hc$ are not a classical control input, e.g. it is not possible to control a human that way, they can be used as a latent signal to change the network inputs and thus act similarly to a velocity control input. As a consequence, the modifiers $\hc$ can be used to change the human prediction in a way that shifts the RNN inputs slightly and, thus, lower the prediction error of the neural network by accounting for task, environmental or collaborative constraints.

\subsubsection{Neural Network Training}
We train the uncontrolled RNN ($\hc=\mathbf{0}$) on the observation data $\data$ in order to find suitable weights $\bm{\theta}$ for the RNN. A batch $\batch\subset\data$ of fixed-length motion trajectories is sampled from the dataset $\data$ at random. We use data augmentation by using a sliding window of size one to obtain all possible trajectories from the data. 

Additionally we randomize the base rotation $r_{\text{base}}$ of the human. The trajectories are splitted into two parts, the first part is fed into the RNN encoder, the second part is used as ground truth in the loss computation.
The training loss is a sum of translational and rotational components:
\begin{align}
  \mathcal{L}(\bm{\theta}) &= ||p_{\text{base}}'-p_{\text{base}}||^2 + ||r'-r||_1
\end{align}
Where we use a mean squared error loss between true and predicted base positions and mean absolute error for the 6d rotation representations $r_{\text{joints}}$ and $r_{\text{base}}$.

The weights $\bm{\theta}$ of the layers are initialized randomly. We use the Adam optimizer~\cite{kingma2014adam} with a learning rate of 0.0001 to train the network.\footnote{Here we optimize with respect to model parameters $\bm{\theta}$, which is not the same as model input used online $\ac$, i.e.,
$\frac{\partial{\mathcal{L}}}{\partial{\bm{\theta}}} \neq \frac{\partial{\mathcal{L}}}{\partial{\ac}}$}

\section{Empirical Evaluation}
\label{sec:experiments}
\begin{figure}
\centering
\subfloat{\includegraphics[width=.4\columnwidth]{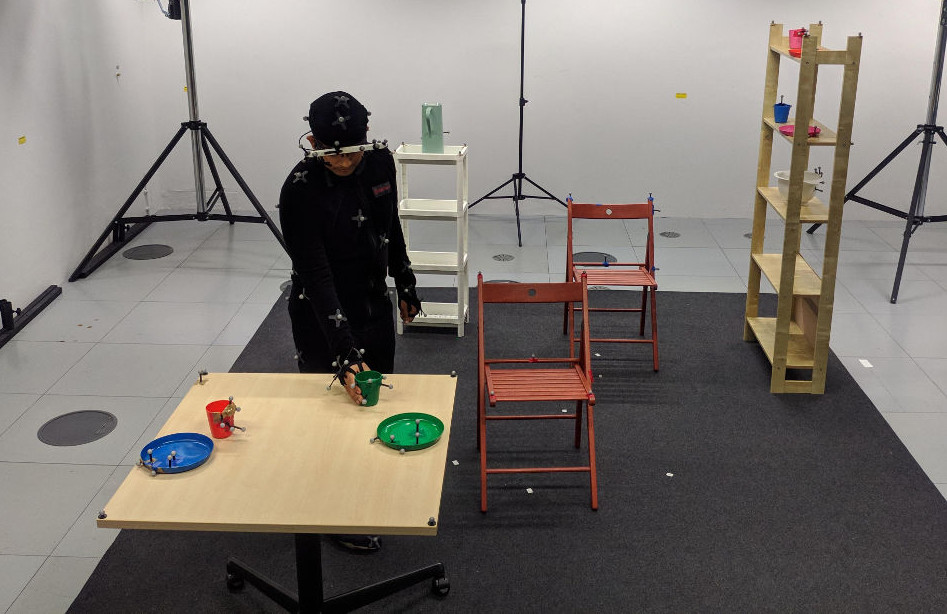}}
\subfloat{\includegraphics[width=.5\columnwidth]{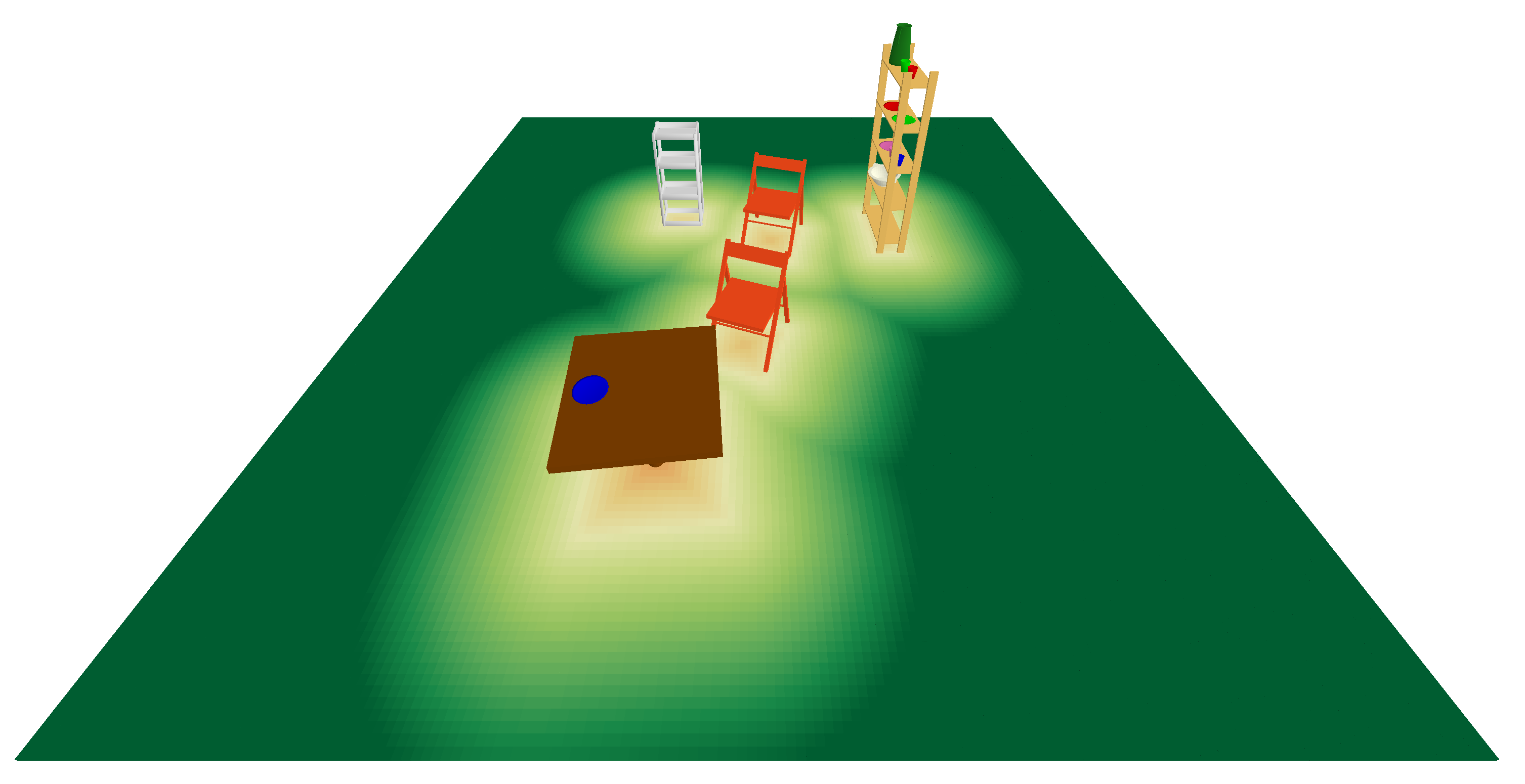}}
\caption{Left: Scene from MoGaze dataset. Right: Signed distance field.}
\label{fig:scene}
\end{figure}
\subsection{Dataset}
To evaluate our framework we use the MoGaze dataset~\cite{kratzer2020mogaze}. MoGaze is a dataset combining full-body motion data and eye-gaze data. In contrast to most other datasets, it contains long, cohered data sequences of object manipulation tasks, such as setting up a table for a fixed number of persons. 

In total it contains 1627 pick and place actions in 180 minutes of motion data with seven recorded participants. This makes the dataset well suited for our application. The data has 120 frames per second which we downsample to 20 frames per second. An example scene of the MoGaze dataset can be seen in Figure~\ref{fig:scene} (left).

\subsection{Human Model}
\begin{table}
  \caption[Joints of the Human Kinematic Model]{The joints of the kinematic model of the human.}
	\centering
	\begin{tabular}{llllll}
          base &  pelvis & torso &  neck & head &  rToe   \\
          lElbow & lWrist & rinnerShoulder & rShoulder &  rElbow & rWrist \\
          lKnee & lAnkle & lToe & rHip & rKnee & rAnkle  \\
          lShoulder & lHip & linnerShoulder &&&\\
        \end{tabular}
        \label{tab:humanmodel}
      \end{table}
      
The human data is modeled using a simple kinematic model with 21 joints. The joints can be seen in Table~\ref{tab:humanmodel}. All joints have a 3D translation and a 3D orientation. Except for the base joint, translations are kept fixed and are only used for adapting the size to different humans, for example, by scaling the length of the forearm.

In total the human data has 66 degrees of freedom (DoFs), 63 DoFs for rotation of the 21 joints and three more DoFs for the base translation.

\subsection{Architectures and Baselines}

For evaluating the prediction performance of our human model we compare with the following architectures:
\begin{itemize}
\item \textbf{zerovel}: The zero-velocity prediction baseline simply predicts no movement. It has been shown to outperform several prediction methods~\cite{martinez2017human}.
\item \textbf{basic-GRU}: A basic RNN with stacked GRU layers. It is similar to the network proposed in~\cite{fragkiadaki2015recurrent}.
\item \textbf{residual}: A RNN with stacked GRU layers and a residual connection, so that individual cells output velocities. It is similar to the network proposed in~\cite{martinez2017human}.
\item \textbf{PVRED}: Our method as described in Section~\ref{sssec:uhd}. It is similar to the network proposed in~\cite{wang2021pvred}.
\end{itemize}

For all architectures we made changes to the original proposed works to improve the prediction performance, such as, changing the rotations to a 6D representation and using the according loss. To further ensure fair comparison, we performed the hyperparameter described in Section~\ref{ssec:hypersearch}.

For evaluating the performance of our framework, we compare the following baselines:
\begin{itemize}
\item \textbf{initial}: We just use the initial (blackbox) prediction outputted by the corresponding architecture.
\item \textbf{sample}: We sample 100 human predictions by adding Gaussian noise to the encoder hidden state $\hidden_t$ before inputting it to the decoder. The predictions are ranked based on a heuristic. The robot is optimized with respect to the samples till a successful trajectory is found (See algorithm~\ref{alg:sample}).
\item \textbf{optim}: We optimize the prediction using our framework. For \textit{basic-GRU} and \textit{residual} only the state input to the RNN is changed, as there are no velocity inputs. For \textit{zerovel} we add latent space modifiers directly to the joint states, leading to a velocity-optimized kinematic solution.
\end{itemize}
 Unless specified otherwise, we use our \textit{PVRED} architecture for the \textit{initial} and \textit{sample} baselines. The method \textit{PVRED-optim} is our full proposed method.

\begin{algorithm}[tb]
\caption{Sample Baseline.}
\begin{algorithmic}
\STATE
\STATE \textbf{Input:} Observed trajectory $\hj_{0:k}$, variance $\sigma^2$
\STATE $\hidden_k \gets \textsc{encoder}(\hj_{0:k})$
\FOR{$i \in \{ 0, 1, 2, \ldots, 100 \}$} 
\STATE $\hidden_k' \gets \hidden_k + \mathcal{N}(\mathbf{0}, \sigma^2)$
\STATE $\text{predictions}[i] \gets \textsc{decoder}(\hidden_k')$
\ENDFOR
\STATE \text{predictions\_ranked} $\gets \textsc{Rank}(\text{predictions})$
\REPEAT
\STATE $\hj_{k:T} \gets \textsc{pop}(\text{predictions\_ranked})$
\STATE $\rj_{k:T},\rc_{k:T},\text{success} \gets$ Trajectory optimization w.r.t. $\hj_{k:T}$
\UNTIL success = TRUE or predictions\_ranked $= \emptyset$
\RETURN $\hj_{k:T},~\rj_{k:T},~\rc_{k:T},~\text{success}$
\end{algorithmic}
\label{alg:sample}
\end{algorithm}

 \subsection{Training Sets and Hyperparameter Search}
 \label{ssec:hypersearch}

We use the whole data of participant 3 for validation. From the remaining participants, we do a hyperparameter search by using 80\% of each participant for training and 20\% for testing.
By excluding a whole participant from training and testing we ensure that the final results reflect behavior for a beforehand unseen human.

During hyperparameter search we change the following parameters:
 \begin{itemize}
 \item Batch size: $|\batch|\in(8, 32, 128)$
 \item Number of GRU layers: $l \in (1, 2, 3)$
 \item Number of hidden units per layer: $d_\hidden \in (100, 200, 300)$
 \end{itemize}
 As weights are initialized randomly and the training behavior depends on that initialization we train each configuration with three different weight initializations. 
 
This leads to 81 trained models per architecture and 243 trained models total. To avoid overfitting, we add dropout and recurrent dropout of 0.2 to the GRU units during training. Each model is trained for 30 epochs, weights are stored after each epoch and the model is evaluated on the test set. For each trajectory we use 1 second (20 timeframes) as input to the networks and 1 second (20 timeframes) as output for loss computation.

 In our following experiments, the model, which performs best on the test set, is used.
 
\subsection{3D rotation representation}

 In a pre-study we compared the exponential map and the 6D rotation representation by training a fixed architecture with both representations. We trained 10 models in total, with 5 randomized weight initializations for each rotation representation. We found that all trained models with 6D representation outperform all the models with exponential map representation in terms of angular error. As a consequence, we use the 6D rotation representation in all remaining experiments.

 \subsection{Goal Experiments}
  \begin{table*}
  \center
  \caption{Oracle Goal Constraint: Comparison of distance to ground truth for our method and several baselines.}
\begin{tabular}{@{}cccccccccccccccc@{}}\toprule
 & \multicolumn{5}{c}{base pos metric} & \multicolumn{5}{c}{angle metric} & \multicolumn{5}{c}{angle arm metric}\\
\cmidrule(lr){2-6}
\cmidrule(lr){7-11}
\cmidrule(lr){12-16}
seconds & 0.4 & 0.8 & 1.2 & 1.6 & 2.0 & 0.4 & 0.8 & 1.2 & 1.6 & 2.0 & 0.4 & 0.8 & 1.2 & 1.6 & 2.0\\
\midrule
zerovel initial & 0.27 & 0.53 & 0.77 & 0.96 & 1.06 & 0.23 & 0.31 & 0.31 & 0.4 & 0.41 & 0.44 & 0.7 & 0.91 & 1.09 & 1.11\\
basic-GRU initial & 0.2 & 0.25 & 0.3 & 0.41 & 0.5 & 0.29 & 0.33 & 0.36 & 0.38 & 0.38 & 0.51 & 0.74 & 0.8 & 0.73 & 0.7\\
residual initial & 0.1 & 0.2 & 0.29 & 0.38 & 0.46 & \textbf{0.18} & 0.26 & 0.29 & 0.33 & 0.34 & \textbf{0.41} & 0.66 & 0.82 & 0.74 & 0.79\\
PVRED initial & \textbf{0.09} & \textbf{0.18} & 0.25 & 0.33 & 0.41 & \textbf{0.18} & \textbf{0.25} & 0.29 & 0.32 & 0.33 & 0.43 & 0.68 & 0.83 & 0.78 & 0.81\\
\midrule
basic-GRU sample & 0.51 & 0.49 & 0.41 & 0.37 & 0.39 & 0.4 & 0.41 & 0.42 & 0.44 & 0.44 & 0.82 & 0.91 & 0.94 & 0.85 & 0.81\\
residual sample & 0.16 & 0.27 & 0.32 & 0.32 & 0.35 & 0.25 & 0.33 & 0.35 & 0.37 & 0.38 & 0.62 & 0.82 & 0.94 & 0.85 & 0.88\\
PVRED sample & 0.15 & 0.25 & 0.29 & 0.28 & 0.31 & 0.26 & 0.32 & 0.34 & 0.35 & 0.36 & 0.57 & 0.82 & 0.86 & 0.83 & 0.82\\
\midrule
zerovel optim & 0.17 & 0.29 & 0.37 & 0.41 & 0.43 & 0.24 & 0.32 & 0.32 & 0.41 & 0.42 & 0.45 & 0.7 & 0.95 & 1.13 & 1.16\\
basic-GRU optim & 0.22 & 0.28 & 0.28 & 0.33 & 0.4 & 0.29 & 0.33 & 0.34 & 0.37 & 0.37 & 0.51 & 0.72 & \textbf{0.79} & 0.74 & 0.7\\
residual optim & 0.1 & 0.19 & 0.24 & 0.24 & 0.27 & 0.19 & 0.26 & \textbf{0.28} & 0.3 & \textbf{0.3} & \textbf{0.41} & \textbf{0.65} & 0.8 & 0.72 & 0.7\\
PVRED optim & \textbf{0.09} & \textbf{0.18} & \textbf{0.23} & \textbf{0.23} & \textbf{0.25} & \textbf{0.18} & 0.26 & \textbf{0.28} & \textbf{0.29} & \textbf{0.3} & \textbf{0.41} & 0.68 & 0.81 & \textbf{0.7} & \textbf{0.68}\\
\bottomrule
\end{tabular}
\label{tab:oracle_goal}
\end{table*}

First, we want to evaluate the performance of our framework for predicting human motion alone. Our hypothesis is that our framework can significantly reduce the distance to ground truth, when a goal constraint obtained from ground truth is added to the hand of the human at the last timestep.

We extract 192 reaching motions from the validation data set. We use trajectories of a duration of 3 seconds, 1 second is used as input to the network and 2 seconds are used for comparison to the prediction. When using our framework, we set a goal constraint to the wrist of the human, which is extracted from the ground truth, on the last timestep. For the sample baselines, we simply take the sample which ends closest to the goal, as we do not consider to optimize robot motion in this experiment.

Table~\ref{tab:oracle_goal} shows the mean over trajectories for different metrics to the ground truth at several timesteps. The \textit{base pos metric} is the distance to the base in meters. The angle metric is the mean relative angle, $L = 2 \; \text{arccos}(\bm{q}^\top \bm{q})$ with $\bm{q}$ being rotations in quaternion representation,  of all joints. The angle arm metric is the same but only for the right arm joints \textit{rElbow} and \textit{rShoulder}.

It can be seen that for the \textit{initial} baselines, \textit{residual} and \textit{PVRED} perform best, with \textit{PVRED} being slightly better in the base position and angle  metric and the residual being slightly better for the angle arm metric. The results of the methods in terms of position at the goal, can be improved by using the \textit{sampling} strategy. However, this increases the angular losses. With our optimization based framework, the results of all architectures improve over the other baselines. While at the beginning of the trajectory, at 0.4 and 0.8 seconds, the pure prediction forecasts have a similar distance to ground truth than our methods, our methods significantly outperform at later timesteps (1.2, 1.6 and 2 seconds), which are closer to the goal.
The best result is achieved by the \textit{PVRED} architecture with our optimization framework.

Our results clearly confirm our experiment's hypothesis and thus demonstrate that our framework, with introducing latent space modifiers in the architecture, can be used to improve motion prediction by specifying constraints and using gradient-based optimization. Note that for prediction purposes, it is possible to obtain a goal location using intent prediction or task knowledge, as we do in~\cite{kratzer2020anticipating}.

\subsection{Joint Collision Avoidance Experiments}
\label{ssec:jcexp}

In this experiment we optimize joint-trajectories ensuring a safety distance between human and robot. We do this by defining an inequality constraint as in Equation~\ref{equ:jointcoll}. We set the distance between the robot and the human base to be at least $d=0.5$m. We also define two equality constraints one for the human and one for the robot goal. For both agents we use a SDF inequality constraint to avoid collision with the scene obstacles (see Figure~\ref{fig:scene}, right). 

To define the planning problems, we extract 76 pick trajectories of human motion from the validation set, with 1sec of observed human motion and 2sec for prediction. We set the goal constraint for the human end position from the ground truth data and we define the start state and goal constraint for
the robot manually, such that the trajectories of the human and robot are likely to interfere.

 \begin{table}
   \caption{Collision experiments with baselines.}
   \label{tab:joint_quant}
   \centering
\begin{tabular}{@{}ccccccc@{}}\toprule
 & \multicolumn{2}{c}{travel dist} &  \multicolumn{3}{c}{smoothness} & \\ 
\cmidrule(lr){2-3}
\cmidrule(lr){4-6}
Method  & human & robot & ms-jerk & ld-jerk & sparc & success rate \\
  \midrule
  with\_coll & 1.28 & 2.28 & -0.03 & -1.62 & -2.51 & 13 \\
  initial & 1.22 & 2.94 & -8.35 & -5.45 & -2.27 & 0 \\
  sample & 1.25 & 3.55 & -15.7 & -5.86 & -2.35 & 3 \\
  human\_avoids & 1.11 & 2.26 & -0.01 & -2.45 & -2.4 & 36 \\
  robot\_avoids & 1.32 & 3.19 & -7.86 & -5.64 & -2.37 & 58 \\
  human\_prio & 1.33 & 2.97 & -13.86 & -6.04 & -2.32 & 58 \\
  ours & 1.32 & 2.72 & -1.26 & -4.66 & -2.29 & 78 \\
  robot\_prio & 1.45 & 2.34 & -0.15 & -3.32 & -2.31 & 83 \\
\bottomrule
\end{tabular}

\end{table}

\subsubsection{Quantitative Results}

We plan coordinated motion trajectories for the human and robot on the set of problem described above using our method, the \textit{initial} baseline, the \textit{sample} baselines by choosing the sample ending closest to the goal, and following additional baselines:
\begin{itemize}
\item \textit{with\_coll} optimizes human and robot without considering the other agent. Both agents can collide. 
\item \textit{human\_avoids} first optimizes a robot trajectory with respect to goal and scene obstacles. Then the human trajectory is optimized to avoid the robot.
\item \textit{robot\_avoids} is similar to \textit{human\_avoids}, but first optimizes a trajectory for the human, then for the robot.
\end{itemize}

We consider 3 variants of our method with different scaling parameters for human and robot in the optimization objective:
\begin{itemize}
\item \textit{human\_prio} variant we use $\alpha^H=100,~\alpha^R=1$
\item \textit{robot\_prio} variant we use $\alpha^H=1,~\alpha^R=100$
\item for the standard variant we use $\alpha^H = \alpha^R=10$.
\end{itemize}

For the resulting trajectories we compute the travel distance of the human and robot base, and the smoothness metrics for the robot trajectory, which  are the negative mean squared jerk, the log dimensional jerk and the spherical arc length. Values closer to zero mean smoother trajectories. 

The median over the 76 trajectories and the success rate can be seen in Table~\ref{tab:joint_quant}. We consider a trajectory as successful, when the distance of the hand of the human to the goal is smaller than $0.1$, the distance of the robot base to the goal is smaller than $0.2$, human and robot do not collide and the optimization objective is smaller than $0.1$.

As the \textit{initial} baseline blindly forecasts, the hand never ends up close to the goal, so that it never found a successful trajectory of human motion. The \textit{sample} baseline only improves this a bit, since it is very unlikely to find collision free trajectories where the human hand ends up close to the goal only by sampling. As both methods output trajectories with the human not reaching to the goal, the median path length of the human base is lower than the one reported by \textit{with\_coll}.

The \textit{robot\_avoids} and \textit{human\_avoids} perform better. However, here the problem is that there is no information flow between the two agents. One of the agents is treated as a blackbox and the other agent tries to find a collision free path, which is often not possible in the timeframe since, for example, the first optimized agent might block the path through a narrow passage.
While the overall success rate is improved over the \textit{initial} and \textit{sample} baselines, the robot often finds very long, non-smooth paths to reach the goal, leading to a long travel distance in the \textit{robot\_avoids} baseline. In the \textit{human\_avoids} baseline, the human is closer to the network prediction and thus often does not reach the goal when the path is blocked by the robot, leading to a lower human base travel distance.

In contrast, our method is able to plan a robot trajectory while adapting the human prediction to the plan, resulting in finding a coordinated motion trajectory for both agents. With a higher parameter for the human costs, as in the \textit{human\_prio} baseline, the changes made to the prediction of the human are too restrictive, so that the prediction often does not fulfill the constraints leading to a lower success rate compared to the \textit{robot\_prio} and \textit{ours} methods.

\textit{Ours} and \textit{robot\_prio} are the best performing methods, with \textit{robot\_prio} increasing the path the human needs to take and decreasing the path length of the robot. Moreover, both methods are amongst the methods with the best smoothness losses for the robot.

\begin{figure}
\centering
\includegraphics[width=\linewidth]{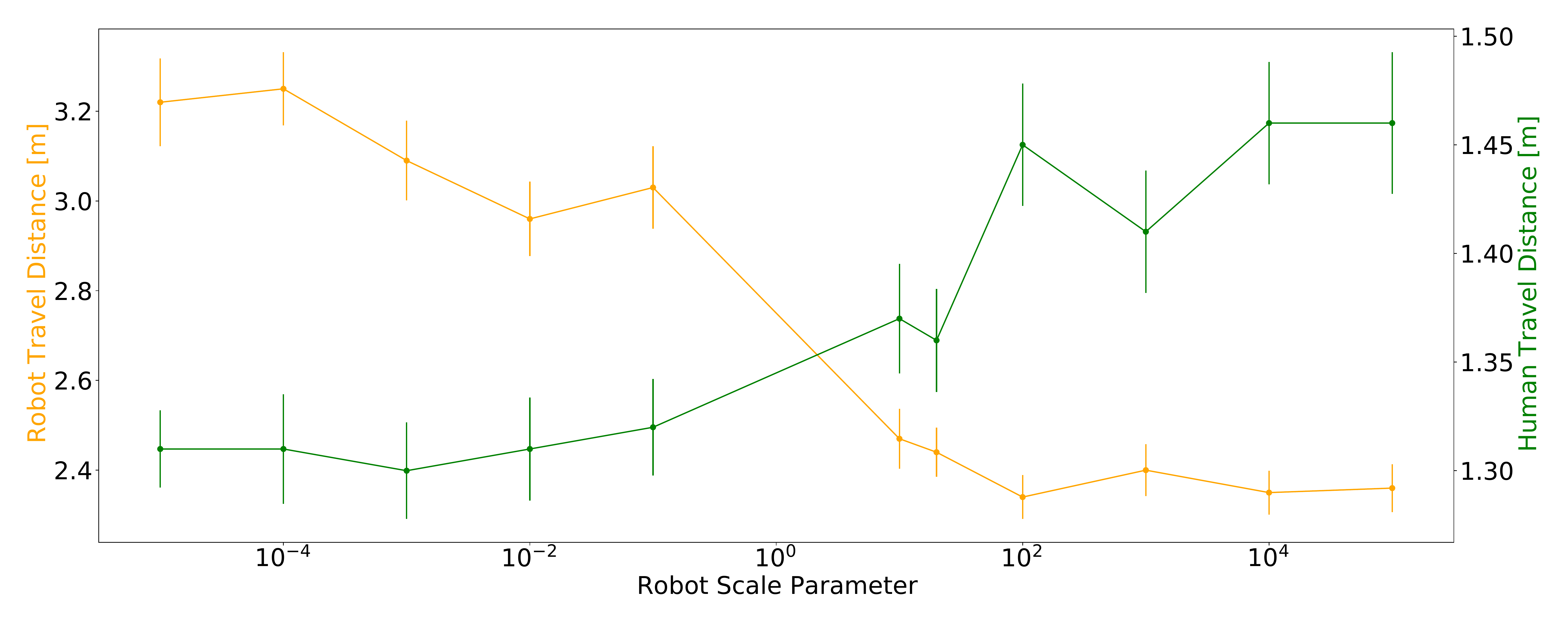}
\caption{Influence of the robot cost scaling parameter. Higher $\alpha^R$ decreases the length of the path of the robot but increases the length of the human path.}
\label{fig:compare_robotparam}
\end{figure}

\subsubsection{Influence of $\alpha^R$}
To further evaluate the influence of the robot cost scaling parameter $\alpha^R$, we run our method multiple times on the trajectories and plot the median path length for human and robot (see Figure~\ref{fig:compare_robotparam}), error bars show the median absolute error. It can be seen that increasing the parameter decreases the robot travel distance but increases the human travel distance. As a consequence, it is possible to use the parameter to balance how much the human and robot should deviate from their predicted and optimal path respectively.

\begin{figure}
  \centering
   \includegraphics[width=.7\columnwidth]{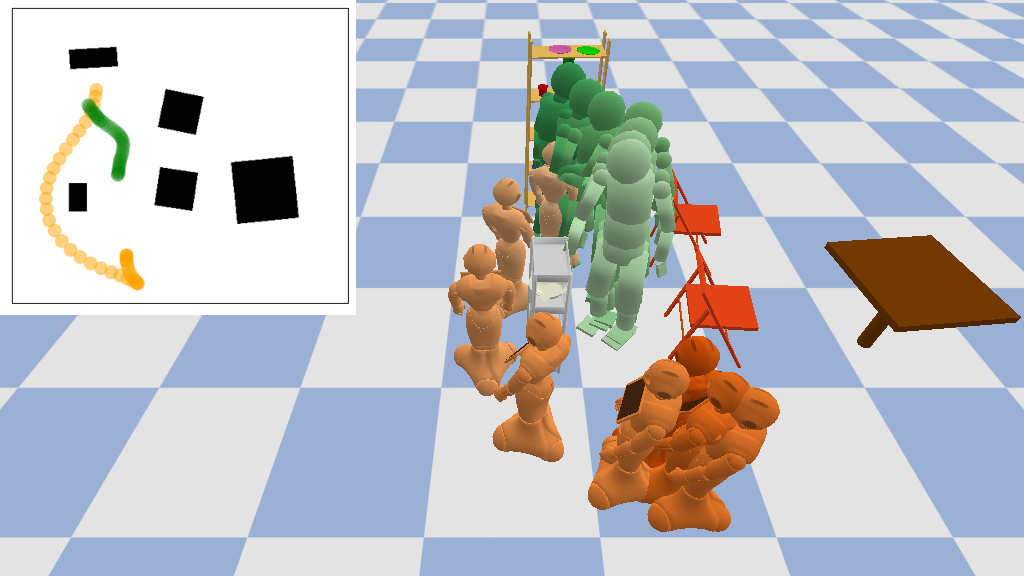}
  \caption{Robot avoiding the human by backing up and taking a longer route.}
  \label{fig:rob_back}
\end{figure}

One can also see that the robot travel distance is changing more than the human's. This is expected since the human is more closely tied to the prediction by the neural network, while the robot can move completely free. For example, the robot can also move backwards to avoid the human (see Figure~\ref{fig:rob_back}). Similar behavior for the human is unlikely since such motions are not included in the dataset. Limitations associated to the dataset will be discussed in section~\ref{sec:discussion}.

 \begin{figure*}
   \centering
   \subfloat{\includegraphics[width=.31\textwidth]{joint_qual/from_left_equ/combined_view.png}}\vspace{0.1cm}
   \subfloat{\includegraphics[width=.31\textwidth]{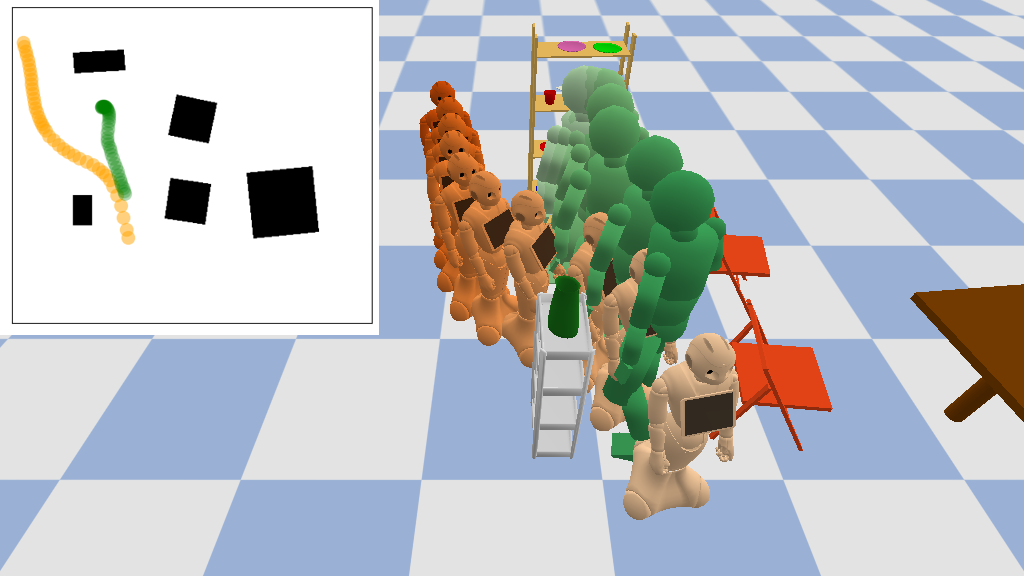}}\vspace{0.1cm}
   \subfloat{\includegraphics[width=.31\textwidth]{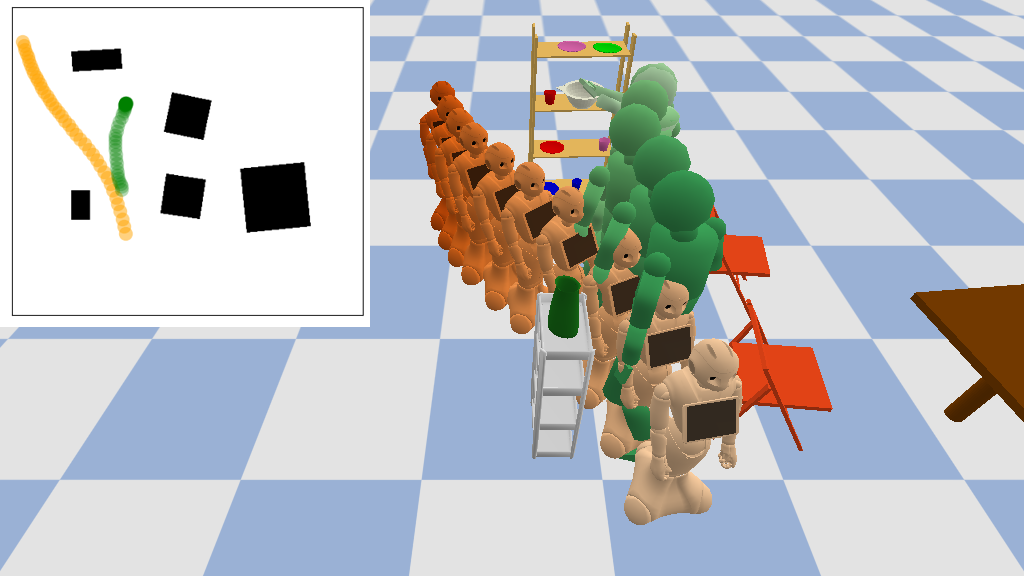}}\\
   \subfloat{\includegraphics[width=.31\textwidth]{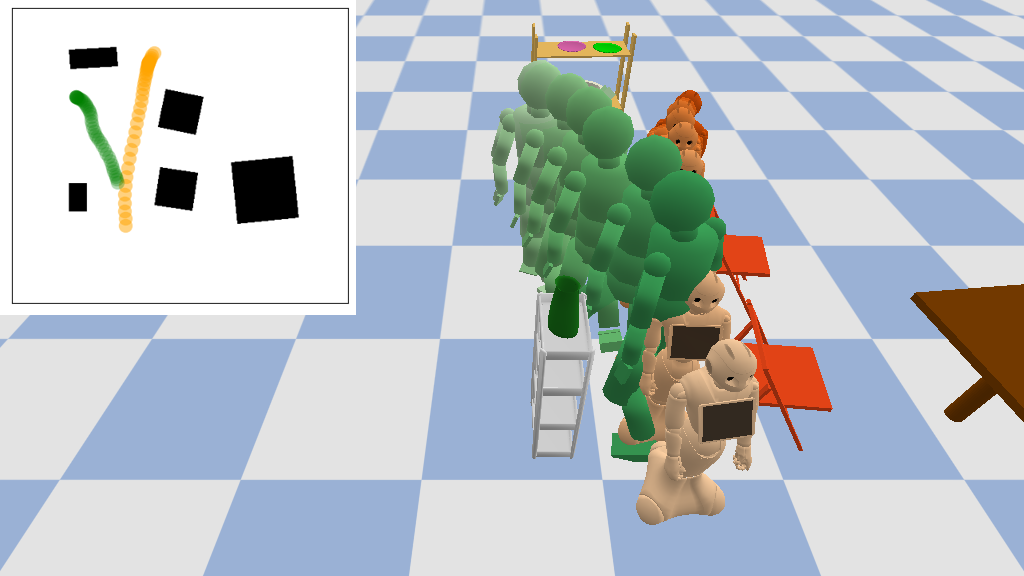}}\vspace{0.1cm}
   \subfloat{\includegraphics[width=.31\textwidth]{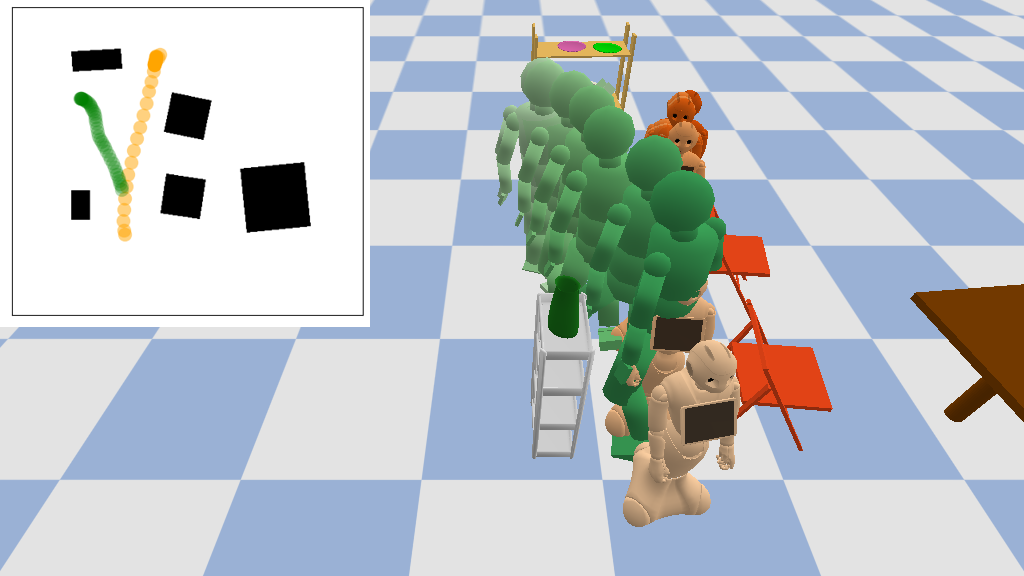}}\vspace{0.1cm}
   \subfloat{\includegraphics[width=.31\textwidth]{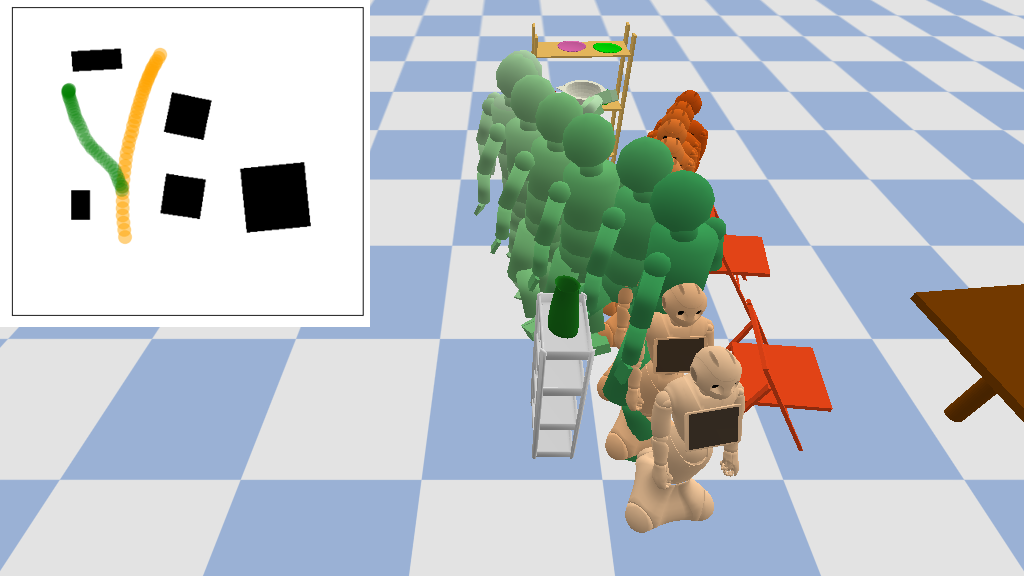}}\\
   \caption{Joint collision avoidance example with robot coming from the left side (top row) and from the right side (bottom row). From left to right: result for our method with $\alpha^H=\alpha^R$, result for our method with human prio $\alpha^H=100,~\alpha^R=1$, result for our method with robot prio $\alpha^H=1,~\alpha^R=100$.}
   \label{fig:joint_qual}
 \end{figure*}

 \subsubsection{Qualitative Results}
 
Figure~\ref{fig:joint_qual} shows example trajectories computed using our method and
the two tuning of the objective function \textit{human\_prio} and \textit{robot\_prio}. The human has the goal to pick up an object from the shelf, the robot has the goal to move to a location on the opposite side. We consider two variants: the robot coming from the left side (top row) and the robot coming from the right side (bottom row). It can be seen that our method adapts to the side the robot is coming from and the human moves away from the robot slightly to make more room for it. 

When $\alpha^H$ is increased (middle column), the human is less adaptive and the robot needs to wait longer till the human passes and speeds up more towards the end of the trajectory. With increased $\alpha^R$ (right column), the human needs to adapt much more and a smoother trajectory for the robot is found.

To conclude, the joint collision avoidance experiments show that our method can be used to plan coordinated trajectories between a human prediction and a robot, and is able to adapt the human prediction to the robot plan. Parameters can be used to influence how much a single agent adapts.

\subsection{Computation Time}

In terms of computation time the bottleneck of our system is computing the gradients for the constraints through the neural network with tensorflow. We use an Intel Core i7 laptop CPU with 2.80GHz for timing experiments.

For computing a 2s trajectory with 20fps we measured the following mean computation times during the Joint Collision experiments: Computing the goal constraint takes 4.97ms and its gradient takes 10.76ms, the SDF collision constraints takes 4.7ms and its gradient takes 10.5ms, the joint collision constraint takes 4.2ms and its gradient takes 9.12ms. 

The optimizer we use usually needs to call the gradient computation once per iteration and the forward pass can be called multiple times. Usually we optimize for 50 to 200 iterations, depending on the used constraints.  As the computation of individual constraints is independent from each other, it would be possible to compute gradients for multiple constraints in parallel, making the cost of 100 iterations of ipopt $\approx$ 2 seconds, which can be further improved by simplifying the network or using better hardware, e.g. GPUs.

\subsection{Handover Experiments}

\begin{figure*}
  \centering
   \subfloat{\includegraphics[width=.31\textwidth]{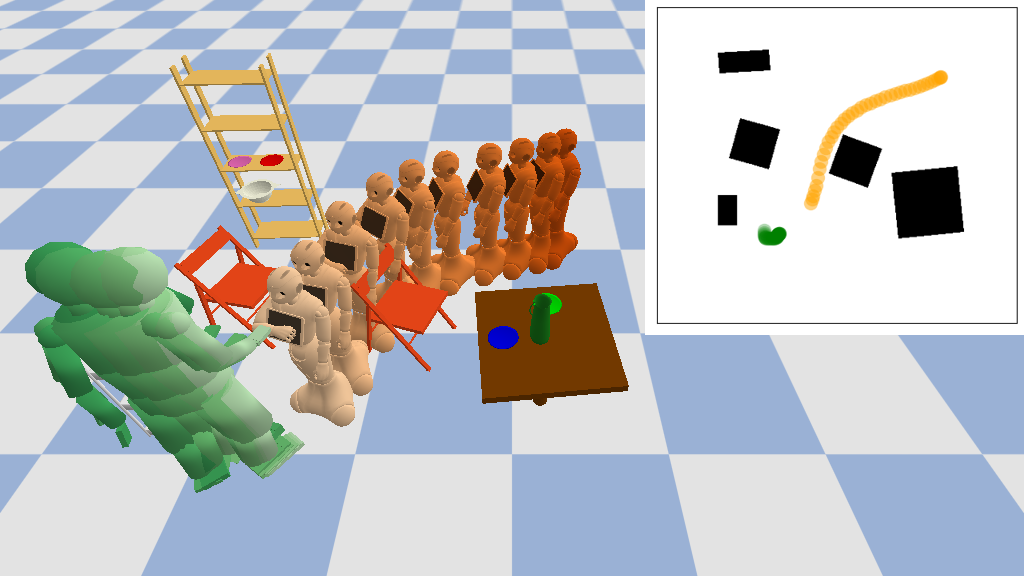}} \vspace{0.1cm}
   \subfloat{\includegraphics[width=.31\textwidth]{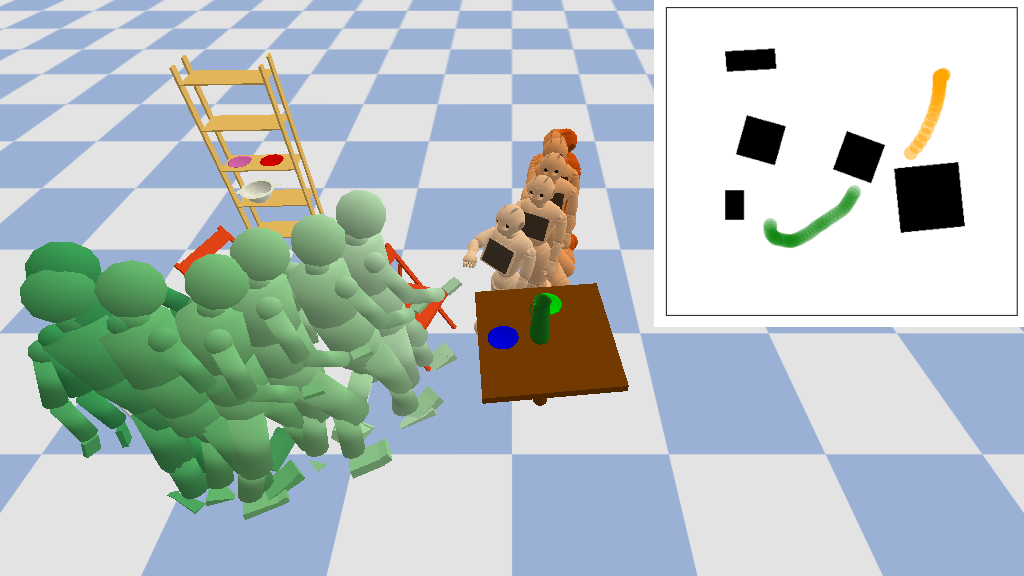}} \vspace{0.1cm}
   \subfloat{\includegraphics[width=.31\textwidth]{handover_qual/ours/combined_view.png}}\\

  \caption{Handover Examples. From left to right: \textit{initial} baseline, \textit{sample} baseline, our method.}
  \label{fig:handover2d}
  \vspace{-0.5cm}
\end{figure*}

We are interested in the planning phase, where the human and robot need to approach each other for
a handover (i.e., within 2 seconds).
We use several trajectories from the dataset and place a robot agent into the scene. We jointly optimize human and robot with obstacle and handover constraints using our framework, the \textit{initial} baseline and the \textit{sample} baseline with ranking the trajectories with respect to their initial handover constraint loss (see Equations~\ref{equ:handover_handpos}, \ref{equ:handover_baserot}).

\subsubsection{Qualitative Results}

Example trajectories are depicted in Figure~\ref{fig:handover2d}, as one can see in the initial prediction (left) the human does not move much, basically the network predicts that the human is standing and turns a bit away from the shelf. The robot has to move all the way towards the human.

With the \textit{sample} baseline (middle) a prediction is chosen, in which the human walks in the direction of the robot. However, in this example, the robot needs to approach too far from the right, in order to perform a handover because it would collide with the chair otherwise. This leads to a trajectory where the robot first slowly rotates before going towards the human, leading to a speedup of the robot at the end of the trajectory.

In contrast, our method (right) is able to make small changes to the human trajectory and directly finds a trajectory where the human and robot move towards each other. As the human trajectory is directly adapted through the optimization,  no handcrafted heuristics are required. The way the human and the robot need to move are better distributed as using the \textit{initial} baseline and the resulting path is more direct and smooth compared to the \textit{sample} baseline.

\begin{table}
  \caption{Handover experiments with baselines.}
  \label{tab:handover_quant}
  \centering
\begin{tabular}{@{}ccccccc@{}}\toprule
 & \multicolumn{2}{c}{travel dist} &  \multicolumn{3}{c}{smoothness} & \\ 
\cmidrule(lr){2-3}
\cmidrule(lr){4-6}
Method  & human & robot & ms-jerk & ld-jerk & sparc & success rate \\
\midrule
initial & 1.2 & 2.19 & -2.57 & -4.6 & -2.11 & 32 \\
sample & 1.34 & 2.16 & -30.34 & -5.17 & -2.15 & 75 \\
ours & 1.13 & 1.09 & -0.09 & -1.89 & -2.19 & 90 \\
\bottomrule
\end{tabular}
\end{table}

\subsubsection{Quantitative Results}
To further evaluate the framework, we extract 100 trajectories from the validation data set and place a robot into the scenario. We then try to find suitable handovers by running our framework as well as the baselines on the trajectories.  Results can be seen in Table~\ref{tab:handover_quant}. We use the same metrics as described in Section~\ref{ssec:jcexp}: path length for human and robot base, smoothness losses and success rates.  Trajectories are considered successful when they not collide with obstacles and reach a handoverloss threshold of $<0.1$ and an objective threshold of $<0.1$.

The prediction of the \textit{initial} baseline does not take any context from the environment or the human position into account. As a consequence it often collides with an obstacle or faces away from the human or towards obstacles. This explains the low success rate of the method and the longer travel distance of the robot, which needs to adapt and take a longer route.

The \textit{sample} baseline performs much better because the method samples multiple futures for the human and thus can select better ones. However, it is still possible that all future trajectories collide with obstacles, for example, when the human walks towards two chairs that are close together. In such scenarios it is hard to find a path in between the chairs just by sampling.

Our method makes it possible to adapt the human future trajectory in a continuous way and can make small changes to the prediction. Thus, the way the travel distance for both agents, human and robot, is smaller compared to the other trajectories and the success rate is significantly improved.

\subsection{Pickup before Handover}
\begin{figure}
  \centering
   \subfloat{\includegraphics[width=0.48\columnwidth]{pbh/std_vid/combined_view}}\vspace{0.1cm}
   \subfloat{\includegraphics[width=0.48\columnwidth]{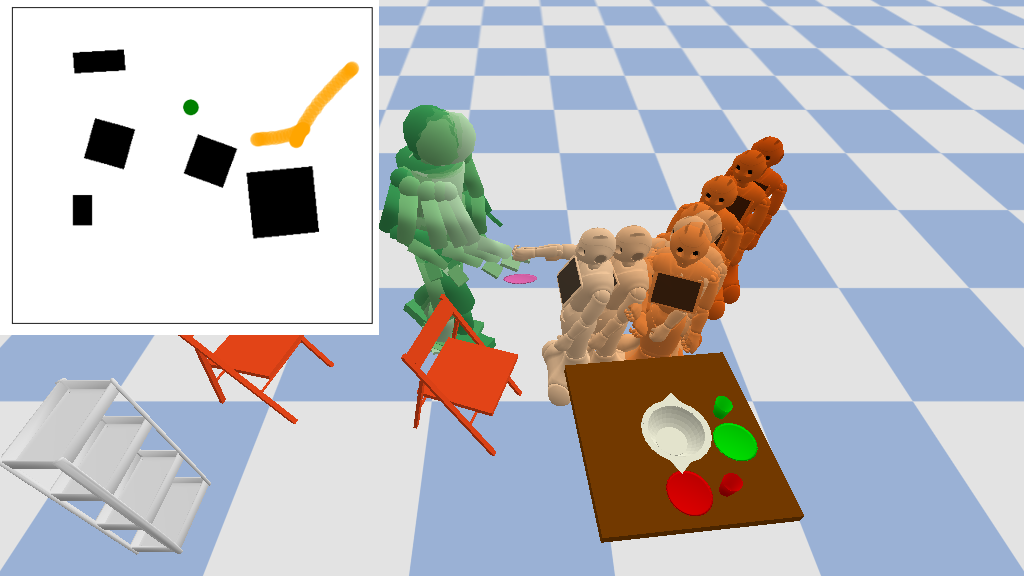}}

  \caption{Example of a joint goal constraint followed by a handover constraint. Left: Human picks up the object and hands it to the robot. Right: Human movement is penalized, hence, the robot picks up the plate.}
  \label{fig:pbh}
\end{figure}
In this experiment we evaluate the \textit{joint goal constraint} (Equation~\ref{equ:jointgoal}). We specify to pickup a plate from the table and do a handover at the end of the trajectory.

An example scenario can be seen in Figure~\ref{fig:pbh}.
The human picks up the object and rotates towards the robot to hand it over (left). We now are interested to see the behavior of our framework when the human is not picking up the plate and thus penalize movement of the human base in the optimization objective. As a result the optimizer outputs a trajectory with the robot picking up the object instead of the human and handing it over to the human (right).
This result shows that our method with the \textit{joint goal constraint} can be used to automatically select a suitable agent for picking up an object.

\section{Discussion and Limitations}
\label{sec:discussion}
\subsection{Dataset}

As shown in our experiments, our system can compute realistic human motions based on the underlying neural network and adapt the motions to account for differentiable constraints by using gradient-based optimization. The neural network is trained on human motion data and the generated motion is optimized to be close to the neural network prediction. 

Therefore the dataset is a crucial part of our framework. Motions that are very different from those in the dataset can not be predicted. For example, in the MoGaze dataset no humans are walking backwards and therefore no backwards walking humans can be predicted. 

An approach to tackle this, is to capture datasets that include a huge range of scenarios, preferably including human-human or human-robot interaction, and thus cover a large fraction of possible human motions. An interesting approach to cover lots of possible human motion trajectories is the AMASS dataset, which aggregates multiple datasets into a common parameterization~\cite{mahmood2019amass}.

\subsection{Trajectory Optimization}

Trajectory optimization generically denotes
Newton's method applied to a trajectory objective.
It is a very powerful class of methods for motion planning and we have
shown that it can also be successfully used to optimize states through a recurrent neural network model. However, trajectory optimization algorithms are local methods, and since
the trajectory objective is typically non-convex,
it can output sub-optimal solutions by getting stuck in poor local minima.
This can be alleviated by random restart,
by combining trajectory optimization with sampling-based approaches
or by convexifing the problem \cite{mainprice2020interior}.

\section{Conclusions}
\label{sec:conclusions}
In this paper we propose a framework for planning coordinated motions in HRC scenarios. The framework uses a recurrent neural network model for predicting human motions and uses gradient-based optimization, to change the predictions while planning a robot trajectory. Thus, it is possible to take environmental, task and coordination constraints into consideration for both: the predicted human trajectory and the planned robot trajectory.

We demonstrated our method on experiments with joint collision avoidance and handovers. Our results show that a recurrent neural network based predictive human model can be used for shared human-robot planning.

For future work we plan to conduct real world experiments with a real robot partner. While our results demonstrate that our method can be used for planning full-body motion trajectories, a challenge in the real world is that the system needs to continuously adapt to the human deviating from the planned motion, which we plan to account for, by iteratively replanning in a model predictive control fashion.

\section*{Acknowledgment}
This work is partially funded by the research alliance ``System Mensch''.
The authors thank the International Max Planck Research School for Intelligent Systems (IMPRS-IS) for supporting Philipp Kratzer.

\bibliographystyle{IEEEtran}
\bibliography{IEEEabrv,21-philipp-coordinated}

\begin{thebibliography}{10}
\providecommand{\url}[1]{#1}
\csname url@samestyle\endcsname
\providecommand{\newblock}{\relax}
\providecommand{\bibinfo}[2]{#2}
\providecommand{\BIBentrySTDinterwordspacing}{\spaceskip=0pt\relax}
\providecommand{\BIBentryALTinterwordstretchfactor}{4}
\providecommand{\BIBentryALTinterwordspacing}{\spaceskip=\fontdimen2\font plus
\BIBentryALTinterwordstretchfactor\fontdimen3\font minus
  \fontdimen4\font\relax}
\providecommand{\BIBforeignlanguage}[2]{{%
\expandafter\ifx\csname l@#1\endcsname\relax
\typeout{** WARNING: IEEEtran.bst: No hyphenation pattern has been}%
\typeout{** loaded for the language `#1'. Using the pattern for}%
\typeout{** the default language instead.}%
\else
\language=\csname l@#1\endcsname
\fi
#2}}
\providecommand{\BIBdecl}{\relax}
\BIBdecl

\bibitem{kratzer2020mogaze}
P.~Kratzer, S.~Bihlmaier, N.~B. Midlagajni, R.~Prakash, M.~Toussaint, and
  J.~Mainprice, ``Mogaze: A dataset of full-body motions that includes
  workspace geometry and eye-gaze,'' \emph{Robotics and Automation Letters},
  vol.~6, no.~2, pp. 367--373, 2020.

\bibitem{martinez2017human}
J.~Martinez, M.~J. Black, and J.~Romero, ``On human motion prediction using
  recurrent neural networks,'' in \emph{Proc. Conf. on Computer Vision and
  Pattern Recognition (CVPR)}, 2017.

\bibitem{pavllo2018quaternet}
D.~Pavllo, D.~Grangier, and M.~Auli, ``Quaternet: A quaternion-based recurrent
  model for human motion,'' in \emph{Proc. Brit. Machine Vision Conf. (BMVC)},
  2018.

\bibitem{wang2021pvred}
H.~Wang, J.~Dong, B.~Cheng, and J.~Feng, ``Pvred: A position-velocity recurrent
  encoder-decoder for human motion prediction,'' \emph{Trans. on Image
  Processing}, vol.~30, pp. 6096--6106, 2021.

\bibitem{kratzer2018}
P.~Kratzer, M.~Toussaint, and J.~Mainprice, ``Towards combining motion
  optimization and data driven dynamical models for human motion prediction,''
  in \emph{Proc. Int. Conf. on Humanoid Robots (Humanoids)}, 2018, pp.
  202--208.

\bibitem{kratzer2020prediction}
------, ``Prediction of human full-body movements with motion optimization and
  recurrent neural networks,'' in \emph{Proc. Int. Conf. on Robotics and
  Automation (ICRA)}, 2020, pp. 1792--1798.

\bibitem{wachter2006implementation}
A.~W{\"a}chter and L.~T. Biegler, ``On the implementation of an interior-point
  filter line-search algorithm for large-scale nonlinear programming,''
  \emph{Mathematical programming}, vol. 106, no.~1, pp. 25--57, 2006.

\bibitem{lasota2017survey}
P.~A. Lasota, T.~Fong, and J.~A. Shah, \emph{A survey of methods for safe
  human-robot interaction}.\hskip 1em plus 0.5em minus 0.4em\relax Now
  Publishers, 2017.

\bibitem{ajoudani2018progress}
A.~Ajoudani, A.~M. Zanchettin, S.~Ivaldi, A.~Albu-Sch{\"a}ffer, K.~Kosuge, and
  O.~Khatib, ``Progress and prospects of the human--robot collaboration,''
  \emph{Auton. Robots}, vol.~42, no.~5, pp. 957--975, 2018.

\bibitem{baraglia2016initiative}
J.~Baraglia, M.~Cakmak, Y.~Nagai, R.~Rao, and M.~Asada, ``Initiative in robot
  assistance during collaborative task execution,'' in \emph{Proc. Int. Conf.
  on Human-Robot Interaction (HRI)}, 2016, pp. 67--74.

\bibitem{schulz2018preferred}
R.~Schulz, P.~Kratzer, and M.~Toussaint, ``Preferred interaction styles for
  human-robot collaboration vary over tasks with different action types,''
  \emph{Frontiers in neurorobotics}, vol.~12, p.~36, 2018.

\bibitem{lasota2015analyzing}
P.~A. Lasota and J.~A. Shah, ``Analyzing the effects of human-aware motion
  planning on close-proximity human--robot collaboration,'' \emph{Human
  factors}, vol.~57, no.~1, pp. 21--33, 2015.

\bibitem{kulic2007pre}
D.~Kuli{\'c} and E.~Croft, ``Pre-collision safety strategies for human-robot
  interaction,'' \emph{Auton. Robots}, vol.~22, no.~2, pp. 149--164, 2007.

\bibitem{Mainprice:13}
J.~Mainprice and D.~Berenson, ``Human-robot collaborative manipulation planning
  using early prediction of human motion,'' in \emph{Proc. Int. Conf. on Intel.
  Robots and Systems (IROS)}, 2013, pp. 299--306.

\bibitem{lasota2014toward}
P.~A. Lasota, G.~F. Rossano, and J.~A. Shah, ``Toward safe close-proximity
  human-robot interaction with standard industrial robots,'' in \emph{Proc.
  Int. Conf. on Automation Science and Engineering (CASE)}, 2014, pp. 339--344.

\bibitem{sisbot2012human}
E.~A. Sisbot and R.~Alami, ``A human-aware manipulation planner,'' \emph{Trans.
  on Robotics}, vol.~28, no.~5, pp. 1045--1057, 2012.

\bibitem{kruse2013human}
T.~Kruse, A.~K. Pandey, R.~Alami, and A.~Kirsch, ``Human-aware robot
  navigation: A survey,'' \emph{Robotics and Auton. Systems}, vol.~61, no.~12,
  pp. 1726--1743, 2013.

\bibitem{bennewitz2005learning}
M.~Bennewitz, W.~Burgard, G.~Cielniak, and S.~Thrun, ``Learning motion patterns
  of people for compliant robot motion,'' \emph{Int. Journal Of Robotic
  Research}, vol.~24, no.~1, pp. 31--48, 2005.

\bibitem{elfring2014learning}
J.~Elfring, R.~Van De~Molengraft, and M.~Steinbuch, ``Learning intentions for
  improved human motion prediction,'' \emph{Robotics and Auton. Systems},
  vol.~62, no.~4, pp. 591--602, 2014.

\bibitem{koppula2013}
H.~S. Koppula, R.~Gupta, and A.~Saxena, ``Learning human activities and object
  affordances from rgb-d videos,'' \emph{Int. Journal Of Robotic Research},
  vol.~32, no.~8, pp. 951--970, 2013.

\bibitem{koppula2016anticipating}
H.~S. Koppula and A.~Saxena, ``Anticipating human activities using object
  affordances for reactive robotic response,'' \emph{Trans. on pattern analysis
  and machine intelligence}, vol.~38, no.~1, pp. 14--29, 2016.

\bibitem{kratzer2020anticipating}
P.~Kratzer, N.~B. Midlagajni, M.~Toussaint, and J.~Mainprice, ``Anticipating
  human intention for full-body motion prediction in object grasping and
  placing tasks,'' in \emph{Proc. Int. Symposium on Robot and Human Interactive
  Communication (RO-MAN)}, 2020, pp. 1157--1163.

\bibitem{le2021hierarchical}
A.~T. Le, P.~Kratzer, S.~Hagenmayer, M.~Toussaint, and J.~Mainprice,
  ``Hierarchical human-motion prediction and logic-geometric programming for
  minimal interference human-robot tasks,'' in \emph{Proc. Int. Symposium on
  Robot and Human Interactive Communication (RO-MAN)}, 2021, pp. 7--14.

\bibitem{rudenko2020human}
A.~Rudenko, L.~Palmieri, M.~Herman, K.~M. Kitani, D.~M. Gavrila, and K.~O.
  Arras, ``Human motion trajectory prediction: A survey,'' \emph{Int. Journal
  Of Robotic Research}, vol.~39, no.~8, pp. 895--935, 2020.

\bibitem{ziebart2009planning}
B.~D. Ziebart, N.~Ratliff, G.~Gallagher, C.~Mertz, K.~Peterson, J.~A. Bagnell,
  M.~Hebert, A.~K. Dey, and S.~Srinivasa, ``Planning-based prediction for
  pedestrians,'' in \emph{Proc. Int. Conf. on Intel. Robots and Systems
  (IROS)}, 2009, pp. 3931--3936.

\bibitem{gupta2018social}
A.~Gupta, J.~Johnson, L.~Fei-Fei, S.~Savarese, and A.~Alahi, ``Social gan:
  Socially acceptable trajectories with generative adversarial networks,'' in
  \emph{Proc. Conf. on Computer Vision and Pattern Recognition (CVPR)}, 2018,
  pp. 2255--2264.

\bibitem{berret2011evidence}
B.~Berret, E.~Chiovetto, F.~Nori, and T.~Pozzo, ``Evidence for composite cost
  functions in arm movement planning: an inverse optimal control approach,''
  \emph{PLoS computational biology}, vol.~7, no.~10, 2011.

\bibitem{mainprice2016goal}
J.~Mainprice, R.~Hayne, and D.~Berenson, ``Goal set inverse optimal control and
  iterative replanning for predicting human reaching motions in shared
  workspaces,'' \emph{Trans. on Robotics}, vol.~32, no.~4, pp. 897--908, 2016.

\bibitem{fragkiadaki2015recurrent}
K.~Fragkiadaki, S.~Levine, P.~Felsen, and J.~Malik, ``Recurrent network models
  for human dynamics,'' in \emph{Proc. Int. Conf. on Computer Vision (ICCV)},
  2015, pp. 4346--4354.

\bibitem{pavllo2019modeling}
D.~Pavllo, C.~Feichtenhofer, M.~Auli, and D.~Grangier, ``Modeling human motion
  with quaternion-based neural networks,'' \emph{Int. Journal of Computer
  Vision}, pp. 1--18, 2019.

\bibitem{li2020dynamic}
M.~Li, S.~Chen, Y.~Zhao, Y.~Zhang, Y.~Wang, and Q.~Tian, ``Dynamic multiscale
  graph neural networks for 3d skeleton based human motion prediction,'' in
  \emph{Proc. Conf. on Computer Vision and Pattern Recognition (CVPR)}, 2020,
  pp. 214--223.

\bibitem{aksan2021spatio}
E.~Aksan, M.~Kaufmann, P.~Cao, and O.~Hilliges, ``A spatio-temporal transformer
  for 3d human motion prediction,'' in \emph{Proc. Int. Conf. on 3D Vision
  (3DV)}.\hskip 1em plus 0.5em minus 0.4em\relax IEEE, 2021, pp. 565--574.

\bibitem{todorov2005generalized}
E.~Todorov and W.~Li, ``A generalized iterative lqg method for locally-optimal
  feedback control of constrained nonlinear stochastic systems,'' in
  \emph{Proc. Amer. Control Conf.}, 2005, pp. 300--306.

\bibitem{ratliff2009chomp}
N.~Ratliff, M.~Zucker, J.~A. Bagnell, and S.~Srinivasa, ``Chomp: Gradient
  optimization techniques for efficient motion planning,'' in \emph{Proc. Int.
  Conf. on Robotics and Automation (ICRA)}, 2009, pp. 489--494.

\bibitem{schulman2013finding}
J.~Schulman, J.~Ho, A.~X. Lee, I.~Awwal, H.~Bradlow, and P.~Abbeel, ``Finding
  locally optimal, collision-free trajectories with sequential convex
  optimization.'' in \emph{Proc. Robotics: science and systems (RSS)}, vol.~9,
  no.~1, 2013, pp. 1--10.

\bibitem{toussaint2014newton}
M.~Toussaint, ``Newton methods for k-order markov constrained motion
  problems,'' \emph{arXiv preprint arXiv:1407.0414}, 2014.

\bibitem{marinho2016functional}
Z.~Marinho, A.~Dragan, A.~Byravan, B.~Boots, S.~Srinivasa, and G.~Gordon,
  ``Functional gradient motion planning in reproducing kernel hilbert spaces,''
  \emph{arXiv preprint arXiv:1601.03648}, 2016.

\bibitem{Toussaint:17}
M.~Toussaint, ``A tutorial on newton methods for constrained trajectory
  optimization and relations to slam, gaussian process smoothing, optimal
  control, and probabilistic inference,'' in \emph{Geometric and numerical
  foundations of movements}.\hskip 1em plus 0.5em minus 0.4em\relax Springer,
  2017, pp. 361--392.

\bibitem{mainprice2020interior}
J.~Mainprice, N.~Ratliff, M.~Toussaint, and S.~Schaal, ``An interior point
  method solving motion planning problems with narrow passages,'' in
  \emph{Proc. Int. Symposium on Robot and Human Interactive Communication
  (RO-MAN)}, 2020.

\bibitem{mordatch2012discovery}
I.~Mordatch, E.~Todorov, and Z.~Popovi{\'c}, ``Discovery of complex behaviors
  through contact-invariant optimization,'' \emph{Trans. on Graphics}, vol.~31,
  no.~4, pp. 1--8, 2012.

\bibitem{mordatch2013animating}
I.~Mordatch, J.~M. Wang, E.~Todorov, and V.~Koltun, ``Animating human lower
  limbs using contact-invariant optimization,'' \emph{Trans. on Graphics},
  vol.~32, no.~6, pp. 1--8, 2013.

\bibitem{fishman2020collaborative}
A.~Fishman, C.~Paxton, W.~Yang, D.~Fox, B.~Boots, and N.~Ratliff,
  ``Collaborative interaction models for optimized human-robot teamwork,'' in
  \emph{Proc. Int. Conf. on Intel. Robots and Systems (IROS)}, 2020, pp.
  11\,221--11\,228.

\bibitem{schaefer2021leveraging}
S.~Schaefer, K.~Leung, B.~Ivanovic, and M.~Pavone, ``Leveraging neural network
  gradients within trajectory optimization for proactive human-robot
  interactions,'' in \emph{Proc. Int. Conf. on Robotics and Automation (ICRA)},
  2021, pp. 9673--9679.

\bibitem{duff2004ma57}
I.~S. Duff, ``Ma57---a code for the solution of sparse symmetric definite and
  indefinite systems,'' \emph{Trans. on Mathematical Software}, vol.~30, no.~2,
  pp. 118--144, 2004.

\bibitem{abadi2016tensorflow}
M.~Abadi, P.~Barham, J.~Chen, Z.~Chen, A.~Davis, J.~Dean, M.~Devin,
  S.~Ghemawat, G.~Irving, M.~Isard \emph{et~al.}, ``Tensorflow: A system for
  large-scale machine learning,'' in \emph{Symposium on operating systems
  design and implementation}, 2016, pp. 265--283.

\bibitem{zhou2019continuity}
Y.~Zhou, C.~Barnes, J.~Lu, J.~Yang, and H.~Li, ``On the continuity of rotation
  representations in neural networks,'' in \emph{Proc. Conf. on Computer Vision
  and Pattern Recognition (CVPR)}, 2019, pp. 5745--5753.

\bibitem{kingma2014adam}
D.~P. Kingma and J.~Ba, ``Adam: A method for stochastic optimization,''
  \emph{arXiv preprint arXiv:1412.6980}, 2014.

\bibitem{mahmood2019amass}
N.~Mahmood, N.~Ghorbani, N.~F. Troje, G.~Pons-Moll, and M.~J. Black, ``Amass:
  Archive of motion capture as surface shapes,'' in \emph{Proc. Int. Conf. on
  Computer Vision (ICCV)}, 2019, pp. 5442--5451.

\end{thebibliography}

\begin{IEEEbiography}
  [{\includegraphics[width=1in,height=1.in,clip,keepaspectratio]{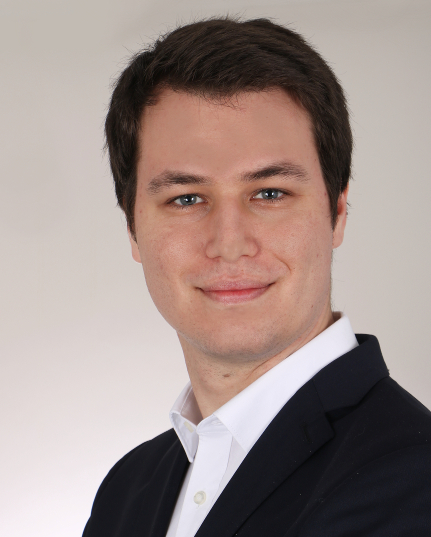}}]{Philipp Kratzer} holds a M. Sc. in Computer Science from University of Stuttgart. Since 2017 he is a doctoral student within the International Max Planck Research School for Intelligent Systems (IMPRS-IS) and researcher at the Machine Learning and Robotics Lab of the University of Stuttgart. His research focuses on improving human-robot interaction by predicting human motion using machine learning techniques.
\end{IEEEbiography}

\vspace{-1cm}

\begin{IEEEbiography}
  [{\includegraphics[width=1in,height=1.in,clip,keepaspectratio]{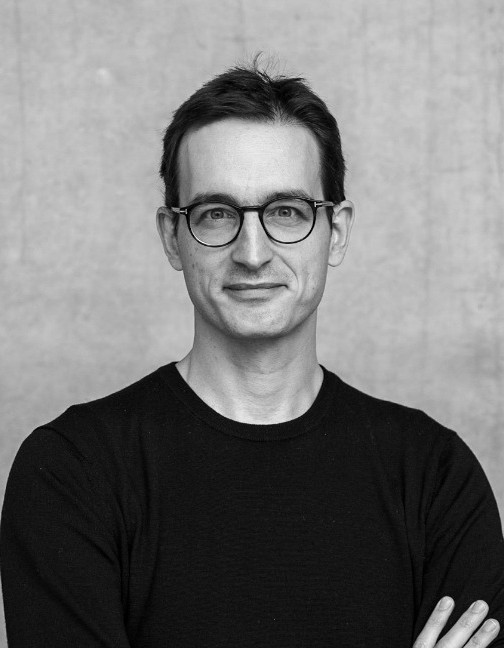}}]{Marc Toussaint} is professor in the area of AI \& Robotics at TU Berlin, lead of the Learning \& Intelligent Systems Lab at the EECS Faculty, and member of the Science Of Intelligence cluster of excellence. From December 2012, he was professor for Machine Learning and Robotics at the University of Stuttgart and Max Planck Fellow at the MPI for Intelligent Systems since November 2018. In 2017/18 he spend a year as visiting scholar at MIT and before that some month at the ML-Robotics lab at Amazon in Berlin. His research bridges between AI planning, machine learning, and robotics, trying to overcome the segregation of data-based AI and model-based AI.
\end{IEEEbiography}

\vspace{-1cm}

\begin{IEEEbiography}
  [{\includegraphics[width=1in,height=1.in,clip,keepaspectratio]{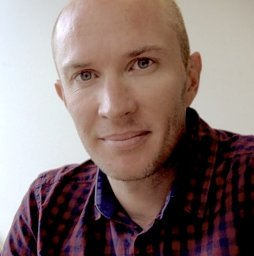}}]{Jim Mainprice}
  is a principal research engineer at Mercedes-Benz since April 2022, where he focuses on autonomous driving. He was previously a substitute professor of computer science at the University of Stuttgart, Germany for two years.  His research interests include robotics, motion planning, motion optimization, and human-motion prediction.
  From January 2013 to December 2014, he was a postdoctoral researcher in the Autonomous Robotic Collaboration Lab at the Worcester Polytechnic Institute (WPI) located in Massachusetts, USA. In January 2015, he moved to the Max Planck Institute for Intelligent Systems in Tübingen, Germany, as a research scientist. From April 2017 to March 2022 he was the founder and leader of the Humans to Robots Motion research group of the University of Stuttgart.
\end{IEEEbiography}

\end{document}